\documentclass{article}

 \usepackage[final]{corl_2026} % Uncomment for the camera-ready ``final'' version.

\usepackage{moreverb,url}
\usepackage{xspace}
\usepackage{xcolor}
\usepackage{float}
\usepackage{tabularx}
\usepackage{array}
\usepackage{amssymb}
\usepackage{amsmath}
\usepackage{amsthm}
\usepackage{amsbsy}
\usepackage{bbm}
\usepackage{dsfont}
\usepackage{hyperref}
\usepackage{tikz}
\usetikzlibrary{fit, positioning}
\usepackage{annotate-equations}

\colorlet{vlacolor}{blue!70!black}
\colorlet{langcolor}{orange!90!black}

\renewcommand{\eqnhighlightshade}{18}
\renewcommand{\eqnannotationfont}{\sffamily\footnotesize}

\usepackage{graphicx}
\usepackage{wrapfig}
\usepackage{booktabs} % for professional tables

\usepackage{subfiles}
\usepackage{csquotes}

\usepackage{array}

\usepackage{graphicx}
\usepackage{subfig} 
\usepackage{multirow}
\newcommand\BibTeX{{\rmfamily B\kern-.05em \textsc{i\kern-.025em b}\kern-.08em
T\kern-.1667em\lower.7ex\hbox{E}\kern-.125emX}}

\definecolor{wine}{RGB}{204, 0, 102}
\definecolor{magenta_wine}{RGB}{158, 44, 143}
\definecolor{dusty_wine}{RGB}{143, 59, 101}
\definecolor{ocean}{RGB}{13, 121, 202}
\definecolor{light_ocean}{RGB}{18, 178, 235}
\definecolor{dark_ocean}{RGB}{10, 89, 148}
\definecolor{grey}{RGB}{170, 170, 170}
\definecolor{light-grey}{RGB}{220, 220, 220}
\definecolor{dark_gray}{rgb}{0.2, 0.2, 0.2} 
\definecolor{med-grey}{rgb}{0.3, 0.3, 0.3} 
\definecolor{grape}{RGB}{112,48,160}
\definecolor{aqua}{RGB}{52,172,139}
\definecolor{dark_aqua}{RGB}{35,115,93}
\definecolor{dark_orange}{RGB}{216,92,0}
\definecolor{vibrant_orange}{RGB}{250, 160, 26}
\definecolor{vibrant_blue}{RGB}{14, 120, 255}
\definecolor{vibrant_pink}{RGB}{255, 0, 104}
\definecolor{dark_red}{RGB}{122, 0, 0}
\definecolor{dark_green}{RGB}{0, 92, 34}
\definecolor{dusty_blue}{RGB}{77, 91, 128}
\definecolor{dark_brown}{RGB}{125, 54, 36}

\newcommand{\para}[1]{\medskip\noindent\textbf{#1. }} 
 
\newcounter{qnum}
\DeclareMathOperator*{\argmax}{argmax}
\DeclareMathOperator*{\argmin}{argmin}

\newcommand{\stateSpace}{\mathcal{S}}
\newcommand{\jstateSpace}{\mathcal{X}}
\newcommand{\actionSpace}{\mathcal{A}}
\newcommand{\obsSpace}{\mathcal{O}}
\newcommand{\markov}{\mathcal{M}}
\newcommand{\globallangspace}{\mathcal{L}}
\newcommand{\locallangset}{\mathcal{L}_{\text{local}}}
\newcommand{\contextSpace}{\mathcal{E}}
\newcommand{\context}{e}

\newcommand{\state}{s}

\newcommand{\traj}{\xi}
\newcommand{\action}{a}

\newcommand{\obs}{o}

\newcommand{\lang}{\ell}
\newcommand{\taskdesc}[1][]{\lang^{\textsf{task}#1}}

\definecolor{basevlacolor}{HTML}{7F7F7F}
\definecolor{vlasftcolor}{HTML}{43928E}
\definecolor{basevlmcolor}{HTML}{BB2867}
\definecolor{sftpolicycolor}{HTML}{6B1AA6}
\definecolor{langpolicycolor}{HTML}{1F77B4}
\definecolor{langpolicycalcolor}{HTML}{FF7F0E}
\definecolor{langpolicynrcolor}{HTML}{BE312C}

\newcommand{\vla}{\textcolor{basevlacolor}{\boldsymbol{\pi}^{\textsf{VLA}}}}
\newcommand{\vlasft}{\textcolor{vlasftcolor}{\boldsymbol{\pi}^{\textsf{VLA-SFT}}}}
\newcommand{\basevlm}{\textcolor{basevlmcolor}{\boldsymbol{\pi}^{\textsf{Base}}}}
\newcommand{\sftpolicy}{\textcolor{sftpolicycolor}{\boldsymbol{\pi}^{\textsf{SFT}}}}
\newcommand{\langpolicy}{\textcolor{langpolicycolor}{\boldsymbol{\pi}^{\textsf{RFT}}}}
\newcommand{\langpolicycal}{\textcolor{langpolicycalcolor}{\boldsymbol{\pi}^{\textsf{CP}}}}
\newcommand{\langpolicynr}{\textcolor{langpolicynrcolor}{\boldsymbol{\pi}^{\textsf{NR}}}}

\newcommand{\langpolicyraw}{\pi^{\textsf{RFT}}}
\newcommand{\vlaraw}{\pi^{\textsf{VLA}}}
\newcommand{\sftpolicyraw}{\pi^{\textsf{SFT}}}

\newcommand{\searchdata}{\mathcal{D}_{\text{search}}}
\newcommand{\rftdata}{\mathcal{D}_{\text{RFT}}}
\newcommand{\advdata}{\mathcal{D}_{\Delta}}

\newcommand{\expdata}{\mathcal{D}_{\text{exp}}}
\newcommand{\nardata}{\mathcal{D}_{\text{nar}}}
\newcommand{\caldata}{\mathcal{D}_{\text{cal}}}

\newcommand{\calibval}{\hat{q}_{\alpha}}

\newcommand{\pip}{\pi_{0.5}}
\newcommand{\pilib}{\pip\text{-LIBERO}}
\newcommand{\pidroid}{\pip\text{-DROID}}

\title{Learning What to Say to Your VLA: Mostly Harmless Vision Language Action Model Steering}
\author{
  Hyun Joe Jeong, Gokul Swamy, Andrea Bajcsy\\
  Robotics Institute\\
  Carnegie Mellon University\\
  \texttt{\{hyunjoej, gswamy, abajcsy\}@andrew.cmu.edu} \\
}

\begin{document}
\maketitle

%===============================================================================

\begin{abstract}
% Vision-Language-Action (VLA) models provide a natural language interface to robot control, but it is often unclear what language inputs will elicit the desired action distribution. 
% Semantically similar instructions can induce drastically different behaviors, while some capabilities may not be recoverable through prompting alone (e.g., because the VLA was not trained on the necessary low-level behavior data). 
% As a result, both human instructions and zero-shot language foundation models can fail to reliably steer VLAs toward successful task execution. 
Vision-Language-Action (VLA) models provide a natural language interface to robot control, but the mapping from language to behavior is often brittle and unintuitive: semantically similar instructions can induce drastically different behaviors, while some capabilities may not be elicitable through prompting alone. 
As a result, both human instructions and zero-shot language models can fail to reliably steer VLAs toward successful task execution.
In this work, we propose a framework that interactively searches for language sequences that improve closed-loop VLA task performance, distills these sequences into a test-time language feedback policy (LFP), and learns an improvement head that predicts when language steering will improve performance. 
 We conformalize this improvement head to prevent harmful steering interventions, where the LFP decreases task performance relative to the original instruction on out-of-distribution scenarios. Crucially, our approach operates on arbitrary frozen pre-trained VLAs, requiring neither access to the original training distribution nor fine-tuning of the underlying model. On seen environments, our conformalized LFP improves base VLA performance by 24.7\% in simulation and 65.0\% in hardware. 
 On visual and semantic perturbations with calibration samples, our conformalized LFP has strong harmlessness guarantees, and produces recovery behaviors not observed with open-loop prompting. 
 Videos of our results can be found on the \href{https://hyunjoe.xyz/LanguagePolicy}{project website}.
\end{abstract}


\section{Introduction}

Vision-Language-Action (VLA) models, learned from large observation-language-action datasets, enable natural language to serve as a flexible test-time interface for eliciting diverse behaviors from a single pre-trained model. Natural language is appealing because it operates at a higher level of abstraction than low-level actions: rather than specifying motor commands directly, a user can control the robot by describing goals, object relations, subgoals, and recovery strategies. Thus, if the frozen VLA already contains useful low-level behaviors, changing the language input can elicit and compose those behaviors without updating the underlying policy.

Although promising, a VLA model's language-to-action mapping is often brittle and unintuitive, making it unpredictable whether a given instruction will reliably elicit the desired sequence of behaviors from the pretrained model. For example, consider a VLA that fails to place all objects into their respective bins when asked to ``put the bowls away.'' Rephrasing the instruction can help the model succeed, but it is difficult to know a priori how to modify the language: some reformulations may succeed, while others may fail despite being semantically similar. This challenge is compounded by the open-vocabulary nature of language: the space of possible utterances is combinatorially large, making exhaustive search over prompts or feedback sequences intractable. Moreover, the desired behavior may not be language steerable at all: the VLA may ignore language conditioning \cite{wanna2026limitedlinguisticdiversityembodied,li2026vlasconfinedcapablegeneralizing}, or the required low-level skill may lie outside its behavioral support, such as asking a pick-and-place robot to pour from a cup. As a result, naive prompting with human instructions or zero-shot language models often fails to consistently elicit correct behaviors.

In this work, our key idea is to learn a language feedback policy to steer a frozen VLA by \textit{interactively searching} for language sequences that improve closed-loop performance. To address the size of the language search space, we interact with the base VLA and search in a structured local language space rather than over arbitrary open-vocabulary utterances. We first use narrated videos of robot behavior to obtain a proposal distribution over language sequences, then interactively evaluate trajectory-level perturbations through closed-loop VLA rollouts to discover language edits that actually elicit successful behavior. We then distill these high-improvement edits into a language feedback policy. However, even when interactive search improves in-distribution performance, the learned policy can still produce harmful steering interventions under distribution shift. 
To make language steering robust out of distribution, we train an improvement head that predicts whether steering will improve performance over the base VLA. \textit{We can do this in a sample-efficient manner because predicting whether improvement is possible is often easier than knowing the precise language sequence that will elicit the improvement}. In particular, we calibrate the improvement head with conformal prediction so that, at deployment, the system language-steers only when predicted improvement is reliable; otherwise, it abstains and falls back to the original instruction.

\noindent \textbf{Statement of Contributions.} We propose an interactive learning algorithm for training a closed-loop language feedback policy (LFP) that learns both \textit{how to steer} a frozen VLA and \textit{when not to steer}, falling back to the original instruction. 
Through simulation and hardware experiments with a Franka Emika manipulator, we show: \textit{(i)} interactive language steering induces better sample complexity than directly fine-tuning the VLA by operating in a higher level of abstraction, leading to better generalization across visual perturbations, semantic perturbations, and novel behavior composition tasks, \textit{(ii)} conformalizing the improvement head prevents performance degradation in out of distribution scenarios without degrading in-distribution performance, and \textit{(iii)} closed-loop language feedback elicits recovery behaviors not observed with open-loop prompt-rephrasing strategies.

\begin{figure}[t!]
    \centering
    \includegraphics[width=1\linewidth]{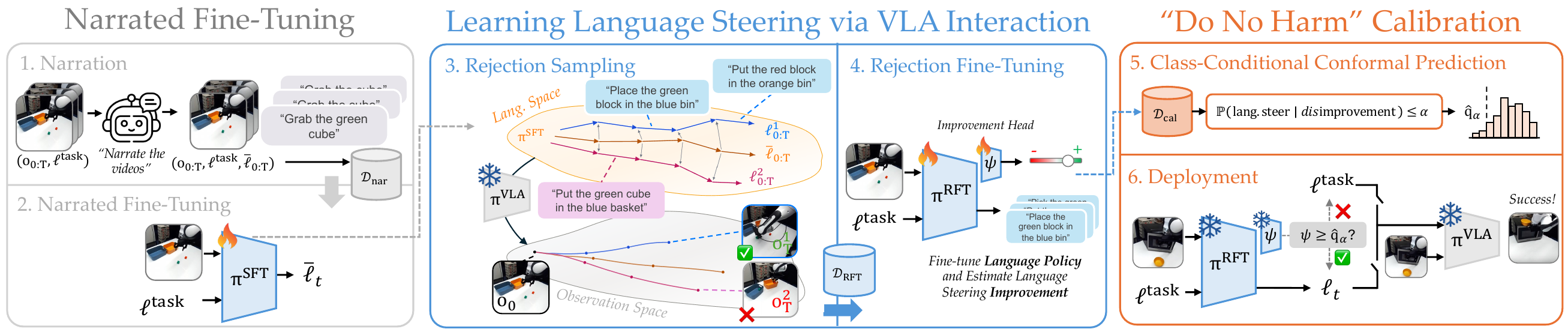}
    \caption{
    \textbf{Overview.} \textit{Left}: robot videos are narrated into per-frame language descriptions, yielding a task-relevant prior without action labels. \textit{Center}: we generate trajectory-level perturbations around this prior, evaluate them  through closed-loop VLA rollouts, and use high-improvement sequences for rejection fine-tuning and improvement-head supervision. \textit{Right}: at deployment, LFP steers the frozen VLA only when the calibrated improvement head predicts benefit; otherwise, the system refuses steering and falls back to the base instruction.
    }
    \label{fig:pipeline}
    \vspace{-1.5em}
\end{figure}
\section{Steering Pre-trained VLA Policies via Natural Language}
\vspace{-1em}
\noindent \textbf{Notation.} Let $t$ index real-world steps, $T$ denote the task horizon, $H$ denote the action-chunk horizon, and let $h\leq H$ denote the number of low-level actions executed before replanning. We consider a frozen, pre-trained VLA model
$\vlaraw(\action_{t:t+H}\mid \obs_t,\state_t,\taskdesc)$ that maps RGB observations
$\obs_t\in\obsSpace$, proprioceptive state $\state_t\in\stateSpace$, and a task
description $\taskdesc\in\globallangspace$ to a distribution over low-level action chunks
$\action_{t:t+H} \in \actionSpace^H$. At execution time, we sample an action chunk from the VLA and execute it in a receding-horizon manner: the robot applies the first $h$ actions from the chunk open-loop, observes the resulting state, and queries the VLA for another length $H$ action chunk.

\noindent \textbf{Language Steering as a Markov Decision Process (MDP).}
We formulate steering a VLA via language as an MDP:
$\markov_{\vlaraw}=(\jstateSpace,\globallangspace,P_{\vlaraw},r)$. States $x_t \in \jstateSpace$ consist of $x_t=(\obs_t,\state_t,\taskdesc)$. Actions $\lang_t \in \globallangspace$ are natural language strings. At each replanning step, we sample a language ``action'' from an LFP $\pi$: $\lang_t\sim\pi(\cdot\mid\obs_t;\taskdesc)$. 
We pass this string to the frozen VLA, which then samples an action chunk
$\action_{t:t+H}\sim\vlaraw(\cdot\mid\obs_t,\state_t,\lang_t)$. The dynamics
$P_{\vlaraw}(x_{t+1}\mid x_t,\lang_t)$ are induced by the frozen VLA composed with the real world. For convenience, we use $\vlaraw \circ \pi$ to denote the closed-loop composite policy that first samples $\lang_t\sim\pi(\cdot\mid\obs_t;\taskdesc)$ and then samples $\action_{t:t+H}\sim\vlaraw(\cdot\mid\obs_t,\state_t,\lang_t)$. Denote $\traj=(\obs_0,\obs_1,\ldots,\obs_T)$ as the real-world observation trajectory induced by such a composite policy. Lastly, let $r(\traj;\taskdesc)\in\{0,1\}$ denote the task reward and $\Pi$ the class of language policies.

Our goal in language steering is to solve the above MDP, i.e., compute 
\begin{equation}
\label{eq:idealsearch}
    \pi^{\star}
    \in
    \argmax_{\pi\in\Pi}
    \mathbb{E}_{\xi \sim \vlaraw \circ \pi}
    \left[
    r(\xi;\taskdesc)
    \right].
\end{equation}

% \noindent \textbf{Objective.}
% Our goal is to find a language control policy which maximizes the reward of the language MDP. 
% Let a rollout in the language MDP induce an observation trajectory $\traj=(\obs_0,\obs_1,\ldots,\obs_T)$. Let $r(\traj;\taskdesc)\in\{0,1\}$ denote the binary trajectory reward, and let $\Pi$ denote the class of language policies. The ideal LFP maximizes expected downstream task success:

\noindent \textbf{Challenges in Language Steering,}
Solving Eq.~\ref{eq:idealsearch} is difficult for at least three reasons. First, the combinatorially large open-vocabulary language space $\globallangspace$ makes exhaustive search intractable. Second, and perhaps more concerningly, the mapping from any sequence of language inputs $\lang_{0:T}$ to VLA actions is often counter-intuitive, so we do not know a priori which utterances will elicit the desired behavior.
Third, the desired behavior may not be language steerable at all: the VLA may ignore language conditioning \cite{wanna2026limitedlinguisticdiversityembodied,li2026vlasconfinedcapablegeneralizing}, or the required low-level skill may lie outside its behavioral support, such as asking a pick-and-place robot to pour from a cup. Thus, while interacting with the VLA is fundamentally necessary to effectively solve Eq.~\ref{eq:idealsearch} (reasons 2, 3), we must carefully structure our language search space to avoid an infeasible amount of real-world interaction (reason 1).

\section{Learning Mostly Harmless Language Feedback Policies via Interaction}
% Directly optimizing Eq.~\ref{eq:idealsearch} is hard because searching through the space open-vocabulary language is intractable and the value of a language edit is only revealed through closed-loop execution of the VLA. We therefore approximate the ideal objective with trajectory-level language search. Rather than searching over arbitrary closed-loop policies, we search over finite language sequences sampled from a narrated proposal distribution and evaluate whether they improve the frozen VLA over the original task instruction.
To balance the necessity of real-world interaction with the practical infeasibility of exhaustive search, we propose to search in a \textit{local language space} centered around narrations of robot behavior generated by a VLM (vision language model). We then distill the outputs of this local search procedure into our LFP (akin to expert iteration \cite{anthony2017thinking, sun2018dual, jain2025smoothseaskilledsailor, levine2013guided}), before conformalizing our LFP to ensure we do not unnecessarily steer in out-of-distribution contexts. See Appendix~\ref{app:implementation} for more.

\noindent \textbf{Language Improvement.} Let $\context\in\contextSpace$ denote the environment context (e.g., geometry of the scene, object appearance and configuration, lighting, and robot viewpoint). 
For a task instruction $\taskdesc$ and context $\context$, let $p(\obs_0\mid\taskdesc,\context)$ denote the robot's initial observation distribution (e.g., wrist and third-person RGB camera data). 
For any candidate language sequence $\ell_{0:T}$, we define the \textit{language improvement}, $\Delta$, of this sequence over passing in the original instruction $\taskdesc$ $T$ times as:
\begin{equation}
\label{eq:language_improvement}
\begin{aligned}
\Delta(\taskdesc,\context,\ell_{0:T})
=
&\;
\mathbb{E}_{\obs_0\sim p(\obs_0\mid\taskdesc,\context)}
\mathbb{E}_{\traj\sim(\vlaraw\circ \ell_{0:T})(\obs_0)}
\left[
r(\traj;\taskdesc)
\right] \\
&-
\mathbb{E}_{\obs_0\sim p(\obs_0\mid\taskdesc,\context)}
\mathbb{E}_{\traj\sim\vlaraw(\obs_0,\taskdesc)}
\left[
r(\traj;\taskdesc)
\right].
\end{aligned}
\end{equation}
Here, $\vlaraw\circ \ell_{0:T}$ denotes the closed-loop rollout distribution obtained by conditioning the frozen VLA on the language sequence $\ell_{0:T}$, while the second term evaluates the base VLA conditioned only on the original task instruction $\taskdesc$. Our goal is to find language sequences with large $\Delta$s.

% Language steering is beneficial for a task and context pair when there exists a candidate language sequence that results in a positive language improvement, $\Delta > 0$. Since $\Delta$ cannot be computed exactly, we estimate it with Monte Carlo rollouts during interactive language search. These estimates both guide the selection of language sequences and supervise an improvement head that predicts when steering should be refused at deployment time.

\subsection{A Three-Phase Approach to Mostly Harmless Language Steering}
We instantiate our approach in three phases, shown in Fig.~\ref{fig:pipeline}. First, since exhaustive search over $\globallangspace$ is computationally intractable, we perform \textbf{narrated fine-tuning}, which gives the LFP a structured proposal distribution over task-relevant language sequences rather than searching over arbitrary open-vocabulary utterances. Second, since it can be difficult to tell a priori what language sequences elicit the right behavior from the frozen VLA, we perform \textbf{interactive language search}, which identifies good language sequences for policy improvement and produces estimates of their language improvement. Third, since steering can be harmful in unseen situations, we \textbf{conformalize} the improvement head to control the rate of harmful steering under deployment-time perturbations.

\noindent \textbf{Phase 1: Narrated Fine-Tuning.}
Since the open-vocabulary language space $\globallangspace$ is too large to search exhaustively, we use narration to convert observation-only demonstrations into a structured proposal distribution for tractable local search. 
Given videos of robot behavior (e.g., as provided by expert demonstration or policy rollouts) for different tasks,
% expert demonstrations 
$\expdata=\{(\xi,\taskdesc)\}$, where each visual trajectory $\xi=(\obs_0,\ldots,\obs_T)$ is paired with a task description $\taskdesc$, we ask a VLM to caption what the robot is doing in the frame in the context of the task description, giving us $\ell_{0:T}=(\ell_0,\ldots,\ell_T)$. This yields a narrated dataset $\nardata$ of tuples $(\obs,\taskdesc,\ell)$, where $\ell$ describes the current subtask at observation $\obs$. We then supervised fine-tune the base VLM to obtain $\sftpolicyraw$:
\begin{equation}
\label{eq:narrated_sft}
    \sftpolicyraw
    =
    \argmax_{\pi\in\Pi}
    \mathbb{E}_{(\obs,\taskdesc,\ell)\sim\nardata}
    \left[
    \log \pi(\ell\mid\obs,\taskdesc)
    \right].
\end{equation}
In summary, narrated fine tuning gives us a strong prior over language actions, making the search over $\globallangspace$ more computationally tractable, akin to the role of pretraining in foundation model training.

% gives $\sftpolicyraw$ a structured prior over language actions, which makes subsequent search in $\globallangspace$ local and tractable.

\noindent \textbf{Phase 2: Interactive Language Search.}
After narrated fine-tuning, we feed language sequences from $\sftpolicyraw$ to the frozen base VLA to find language sequences that improve performance over the task description. Given $(\taskdesc,\context)$, $\sftpolicyraw$ generates a \textit{seed sequence} $\bar{\ell}_{0:T}$ by interacting with the VLA and the real world. We then use an LLM to generate $N$ trajectory-level semantic perturbations of this seed sequence, prompting it to  rewrite the entire seed sequences while preserving the task semantics. Practically, these perturbations vary object references, verbs, and compositional phrasing. This process yields a \textit{local language set}
$\locallangset(\bar{\ell}_{0:T})=\{\ell^{(1)}_{0:T},\ldots,\ell^{(N)}_{0:T}\}$, which we scope our search to. We then use rollouts to compute a Monte Carlo estimate of improvement, yielding a $\widehat{\Delta}$ for each language sequence. Finally, we select the language sequence with the highest $\widehat{\Delta}$, i.e.
\begin{equation}
\label{eq:interactive_search}
    \ell^\star_{0:T}(\taskdesc,\context)
    =
    \argmax_{\ell^{(n)}_{0:T}\in\locallangset(\bar{\ell}_{0:T})}
    \widehat{\Delta}(\taskdesc,\context,\ell^{(n)}_{0:T}).
\end{equation}
Next, we collect the selected sequences into a rejection fine-tuning (RFT) dataset $(\obs,\taskdesc,\ell^\star) \in \rftdata$, before distilling them into our LFP (i.e., expert iteration \cite{anthony2017thinking, sun2018dual, jain2025smoothseaskilledsailor, levine2013guided}):
\begin{equation}
    \langpolicyraw
    =
    \argmax_{\pi\in\Pi}
    \mathbb{E}_{(\obs,\taskdesc,\ell^\star)\sim\rftdata}
    \left[
    \log \pi(\ell^\star \mid \obs,\taskdesc)
    \right].
\end{equation}
As we only train on language sequences with positive expected improvement, RFT should only improve performance in training scenarios under standard assumptions. However, we empirically observe that \textit{(i)} the VLA may not be \textit{language steerable} for some tasks, so language steering can be ignored or mapped to unintended action distributions that harm performance relative to the base VLA; and \textit{(ii)} this risk is exacerbated when RFT is performed on a relatively small dataset, which can lead to out-of-distribution performance degradation. In response, we use the same data to train an \textit{improvement head} to predict at which initial states steering is beneficial. We define an improvement dataset 
$(\obs_0,\taskdesc,\context,\widehat{\Delta})\in\advdata$,
where $\widehat{\Delta}$ is the estimated language improvement.
We then train our improvement head on $\advdata$ via square loss regression to predict how beneficial steering is:
\begin{equation}
    \psi
    =
    \argmin_{\psi\in\Psi}
    \mathbb{E}_{(\obs_0,\taskdesc,\context,\widehat{\Delta})\sim\advdata}
    \left[
    \left(
    \psi(\obs_0,\taskdesc)-\widehat{\Delta}
    \right)^2
    \right].
\end{equation}
% We note that $\widehat{\Delta}$ is measured at the task and environment level, deployment is observation-conditioned: $\langpolicyraw$ is queried closed-loop from the current observation, and $\context$ is not provided explicitly. 
We pass $\obs_0$ as input to $\psi$ so that the improvement head can implicitly condition on context $\context$ via $\obs_0$. 

% We steer only if this predicted improvement exceeds a threshold $\calibval$: if $\psi(\obs_0,\taskdesc)\geq\calibval$, we deploy the learned LFP $\langpolicyraw$ and execute the steered closed-loop system; otherwise, we refuse to steer and fall back to the base VLA conditioned only on $\taskdesc$. The natural choice is $\calibval=0$, which steers whenever the improvement head predicts non-negative benefit.

\noindent \textbf{Phase 3: Conformalizing the Improvement Head.}  Perhaps the most natural ``wrapper algorithm'' would be to deploy the learned LFP $\langpolicyraw$ iff $\psi(\obs_0,\taskdesc)\geq0$, falling back to the static task description $\taskdesc$ otherwise. However, when queried in an out-of-distribution scenario, $\psi$ may incorrectly predict the LFP improves performance when it actually reduces task success rates.

Ideally, we'd \textit{do no harm}, only switching to our LFP at initial states where it improves performance over the raw task description. While we could achieve this goal via iteratively collecting data to re-train both our LFP and improvement head, this approach ignores a key asymmetry: \textit{it is easier to predict whether improvement is possible than knowing the precise language sequence to elicit this improved behavior.} Thus, we propose taking a lighter-weight approach: \textit{conformalizing} the improvement head to control its false positive rate (FPR), rather than retraining both components. In particular, we propose using class-conditional conformal prediction \cite{vovk2005algorithmic}. Mechanically, this boils down to choosing a higher steering threshold $\calibval > 0$ on improvement head predictions to limit the chances of us deploying the LFP when repeating the raw task description would have worked better.

% When deployed in a novel setting, the learned improvement head $\psi$ may not reliably predict whether language steering will help or harm. Even if $\langpolicyraw$ was beneficial during training, deploying $\vlaraw$ with $\langpolicyraw$ can degrade task performance when the VLA ignores language interventions or lacks the required skill. We therefore want the steered system to \textit{do no harm}: it should steer only when the predicted improvement is reliable, and otherwise fall back to the base VLA. This amounts to controlling false positives, i.e., cases where the system allows steering even though the empirical language improvement is negative.

% To control this error rate, we use class-conditional conformal prediction \cite{vovk2005algorithmic} to choose a steering threshold $\calibval$ on the learned improvement score. Class-conditional calibration is appropriate because the error we care about is not \textit{overall classification accuracy}, but \textit{harmful steering when the improvement is negative}.

More formally, let $\caldata=\{(X^i,\context^i,\widehat{\Delta}^i)\}_{i=1}^{N_{\text{cal}}}$ denote our calibration set, where $X^i=(\obs^i_0,\taskdesc[,i])$, $\context^i$ is the environment, and $\widehat{\Delta}^i$ is the empirical language improvement. We construct $\caldata$ from held-out visual and semantic OOD perturbations that neither $\langpolicyraw$ nor $\psi$ is trained on. Define $Y^i=\mathbb{I}\{\widehat{\Delta}^i \geq 0\}$. Then, the label $Y^i=0$ implies steering harms performance. Let $s^i=\psi(X^i)$ be the predicted improvement. We set $\calibval$ to the empirical $(1-\alpha)$ quantile of the harmful subset of the calibration dataset, i.e., $\calibval = \operatorname{Quantile}_{1-\alpha}
\left(\{s^i : \{(s^i, Y_i)\}_{i=1}^{N_{\text{cal}}},\ Y_i = 0\}\right)$. Under standard exchangeability assumptions (i.e., that the calibration and test examples are drawn from the same distribution), we prove in Appendix~\ref{app:conformal_proof} that we control the false positive rate of $\psi$ at $\alpha$:
\begin{equation}
\label{eq:cp}
    \mathbb{P}\left(
    \psi(X) \ge \calibval
    \mid Y=0
    \right)
    \le \alpha.
\end{equation}
Then, by steering when $\psi(X)\geq\calibval$, we also control the performance degradation probability at $\alpha$.

% This yields the class-conditional guarantee
% \begin{equation}
% \label{eq:cp}
%     \mathbb{P}\left(
%     \psi(X) \ge \calibval
%     \mid Y=0
%     \right)
%     \le \alpha,
% \end{equation}
% meaning that among examples where steering is empirically harmful, the probability of incorrectly allowing steering is at most $\alpha$.
% The guarantee holds under the standard conformal exchangeability assumption; in our setting, this requires the calibration and deployment examples to be drawn from the same task and environment distribution. The guarantee follows from the standard class-conditional conformal prediction; we provide the proof in Appendix~\ref{app:conformal_proof}.

\section{Experiments}
\label{sec:result}
In simulation and hardware environments, our experiments ask: \textbf{(Q1)} How does language steering compare to the base VLA and direct VLA fine-tuning in terms of in-distribution performance, out-of-distribution performance, and sample complexity? \textbf{(Q2)} Can conformalization prevent harmful steering? \textbf{(Q3)} Is closed-loop language feedback necessary, or is open-loop prompt search sufficient? \textbf{(Q4)} Does language steering elicit desired behaviors on various VLAs? Throughout, we evaluate on a suite of visual scene and semantic task perturbations.
% to test the robustness and generalization of each method.
\begin{figure}
    \centering
    \includegraphics[width=1\linewidth]{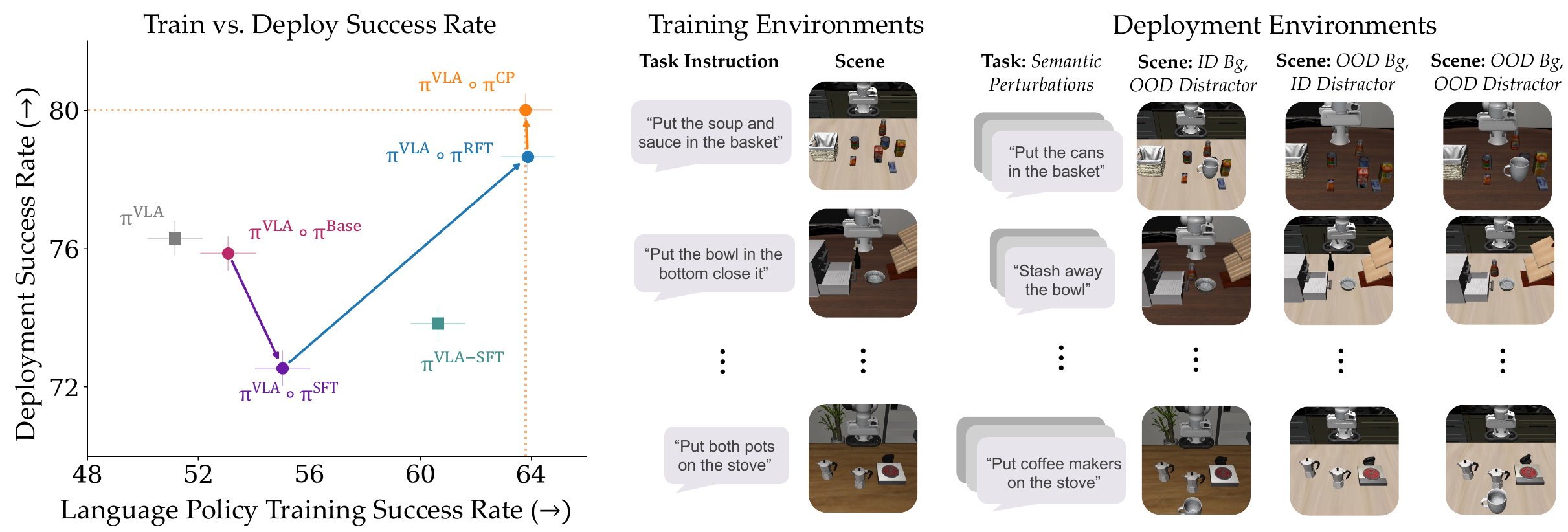}
    \caption{\textbf{Simulation.} \textit{Left}: Baseline comparisons are shown in the language-policy training environment vs. held-out deployment environments; error bars are standard error over pooled episodes across perturbation combinations. $\sftpolicy$ improves ID performance but degrades OOD performance relative to $\basevlm$, showing that narration is not enough for robustness. Interactive search and conformalization recover OOD robustness, with $\langpolicy$ and $\langpolicycal$ outperforming all baselines. Note that absolute success is higher in deployment environments because LIBERO-10 is near-saturated for the base VLA. \textit{Right}: training and deployment environments are visualized for selected tasks.}
    \label{fig:baselines}
    \vspace{-1.5em}
\end{figure}

\subsection{Experimental Setup}
\noindent \textbf{VLA.}
We use $\vla := \pip$ \cite{intelligence2025pi05visionlanguageactionmodelopenworld} fine-tuned on LIBERO ($\pilib$) \cite{liu2023liberobenchmarkingknowledgetransfer} in simulation, DROID ($\pidroid$) in hardware~\cite{khazatsky2025droidlargescaleinthewildrobot}, and run on a Franka robot with wrist and third person cameras.

\noindent \textbf{Language Policy Training.} Our LFP is initialized as the Qwen3-VL-4B-Instruct \cite{bai2025qwen3vltechnicalreport} model and fine-tuned via LoRA \cite{hu2021loralowrankadaptationlarge}. 
At train and test-time, we greedily decode $\lang_t$ from $\langpolicy$. In simulation and hardware, $\expdata$ consists of $K=50$ videos per task rolled out from the respective base VLA policies, which are then narrated by Molmo2-8B \cite{clark2026molmo2openweightsdata} to produce $\nardata$ and train $\sftpolicy$. For interactive search, we generate $N=16$ language sequence perturbations around the narration proposed by $\sftpolicy$ via GPT-5.4 \cite{openai_gpt54thinking_2026}; the resulting dataset $\searchdata$ is used for further training the LFP as well as the improvement head. 
We calibrate the intervention threshold on held-out visual and semantic perturbation episodes where the empirical task-level improvement is negative. More details on datasets, prompts, hyperparameters, and conformalization are available in Appendix~\ref{app:implementation}.

\noindent \textbf{Simulation: Scenarios.}
We use LIBERO-OOD \cite{wu2026saysteeringvisionlanguageactionmodels}, which augments the LIBERO-10 long-horizon manipulation suite with visual and semantic perturbations. 
LIBERO-OOD includes two visual perturbations: \textbf{Visual-Scene}, which adds novel distractor objects, and \textbf{Visual-Viewpoint}, which changes the background and camera viewpoint. 
We further add \textbf{Visual-Scene-Viewpoint}, which adds novel distractors to Visual-Viewpoint, and two more semantic perturbations per task. We define ID/OOD with respect to the LFP $\langpolicy$ training environment. We train $\langpolicy$ on \textbf{Visual-Viewpoint}; this environment is ID for $\langpolicy$, but visually OOD for the original $\pilib$ VLA. 
During evaluation, we run all models on both the LFP training environments and held-out visual environments: \textbf{Visual-Scene-Viewpoint} (ID scene, OOD distractor objects), \textbf{LIBERO-10} (OOD scene, ID distractor objects), and \textbf{Visual-Scene} (OOD scene, OOD distractor objects). 
For each visual environment, we additionally evaluate on 5 semantic perturbations, yielding 200 visual-semantic evaluation combinations in total. Example environments are shown in Fig.~\ref{fig:baselines} (right). Finally, we evaluate novel task composition in \textbf{Compose}, which contains 13 unseen combinations of learned behaviors. See App.~\ref{app:benchmark} for details.

\begin{figure}[t!]
    \centering
    \includegraphics[width=0.96\linewidth]{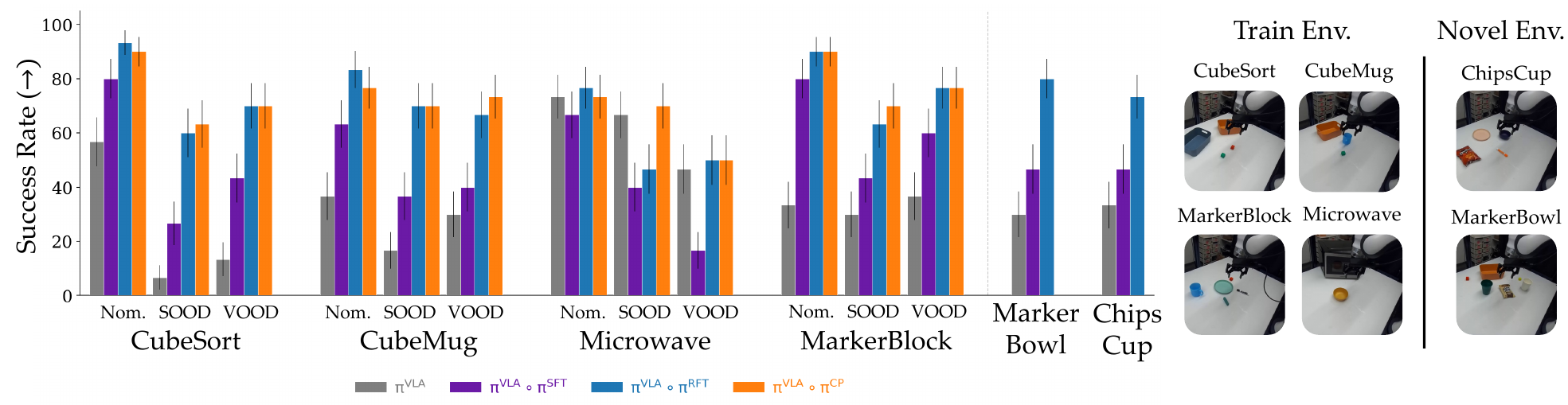}
    \vspace{-0.5em}
    \caption{\textbf{Hardware.} We have four training tasks (Nom.) and their test-time visual (VOOD) and semantic (SOOD) perturbations, and two novel tasks (right). $\langpolicy$ usually improves task performance, and conformalization reduces harmful interventions on tasks such as \textbf{Microwave}.}
    \label{fig:hardwareresults}
    \vspace{-1.5em}
\end{figure}

\noindent \textbf{Hardware: Scenarios.} 
We run $\vla$ zero-shot without fine-tuning. 
For training our LFP, we design four tabletop tasks that include both steerable and un-steerable scenarios, each with visual and semantic perturbations (shown in Fig.~\ref{fig:hardwareresults}). 
\textbf{CubeSort}: robot must place a green cube in the blue basket and a red cube in the orange basket; \textbf{CubeMug}: robot must place a green cube in the blue mug and then place the mug in the orange basket; \textbf{MarkerBlock}: robot must place a marker and a green block on a plate; \textbf{Microwave}: robot must place an orange bowl on top of a microwave and then close the microwave door. Details on prompts and setup in App.~\ref{app:benchmark}.
Visual perturbations introduce additional distractor objects into the scene. 
For evaluation, semantic perturbations consist of one paraphrased instruction per task that preserve the task goal while varying wording, object references, and compositional structure. For example, in \textbf{CubeSort}, the training-time $\taskdesc$ is ``put the green cube in the blue basket and the red cube in the orange basket and put away the green cube and the red cube in similar color baskets'' and a perturbation asks  ``\underline{put away} the green cube and the red cube in \underline{similar color baskets}.''
% These perturbations test robustness to language variation beyond the original training instruction, and 
The full set of semantic perturbations are in Table~\ref{tab:droid_task_descriptions}. 
We additionally test two unseen tasks, \textbf{ChipsCup}, where the robot must put the chip bag and green cup in the basket, and \textbf{MarkerBowl}, where it should put the marker in the blue bowl and then place the bowl on a plate. 
For these unseen tasks, we evaluate zero-shot generalization of our LFP. 
\subsection{Q1: Why Should We Do Language Steering of VLAs?}
We first evaluate whether language steering outperforms the base VLA and VLA fine-tuning across two axes: \textit{(i)} in- and out-of-distribution robustness under visual and semantic perturbations, and \textit{(ii)} sample complexity, measured by varying the amount of successful on-policy fine-tuning data.

\noindent \textbf{Methods.} We compare $\langpolicy$ and $\langpolicycal$ ($\alpha=0.10$) against the base VLA $\vla$, direct VLA fine-tuning $\vlasft$, the off-the-shelf VLM $\basevlm$, and the narrated SFT language policy $\sftpolicy$. For a fair comparison, $\vlasft$ uses the same successful rollouts from the base VLA as our method, but also uses action labels: we fine-tune the VLA with LoRA \cite{hu2021loralowrankadaptationlarge} using the conditional flow matching loss on action chunks \cite{intelligence2025pi05visionlanguageactionmodelopenworld}.
\vspace{-1em}

\begin{wrapfigure}{r}{0.25\columnwidth}
    %\vspace{-1.2em}
    \centering
    \includegraphics[width=\linewidth]{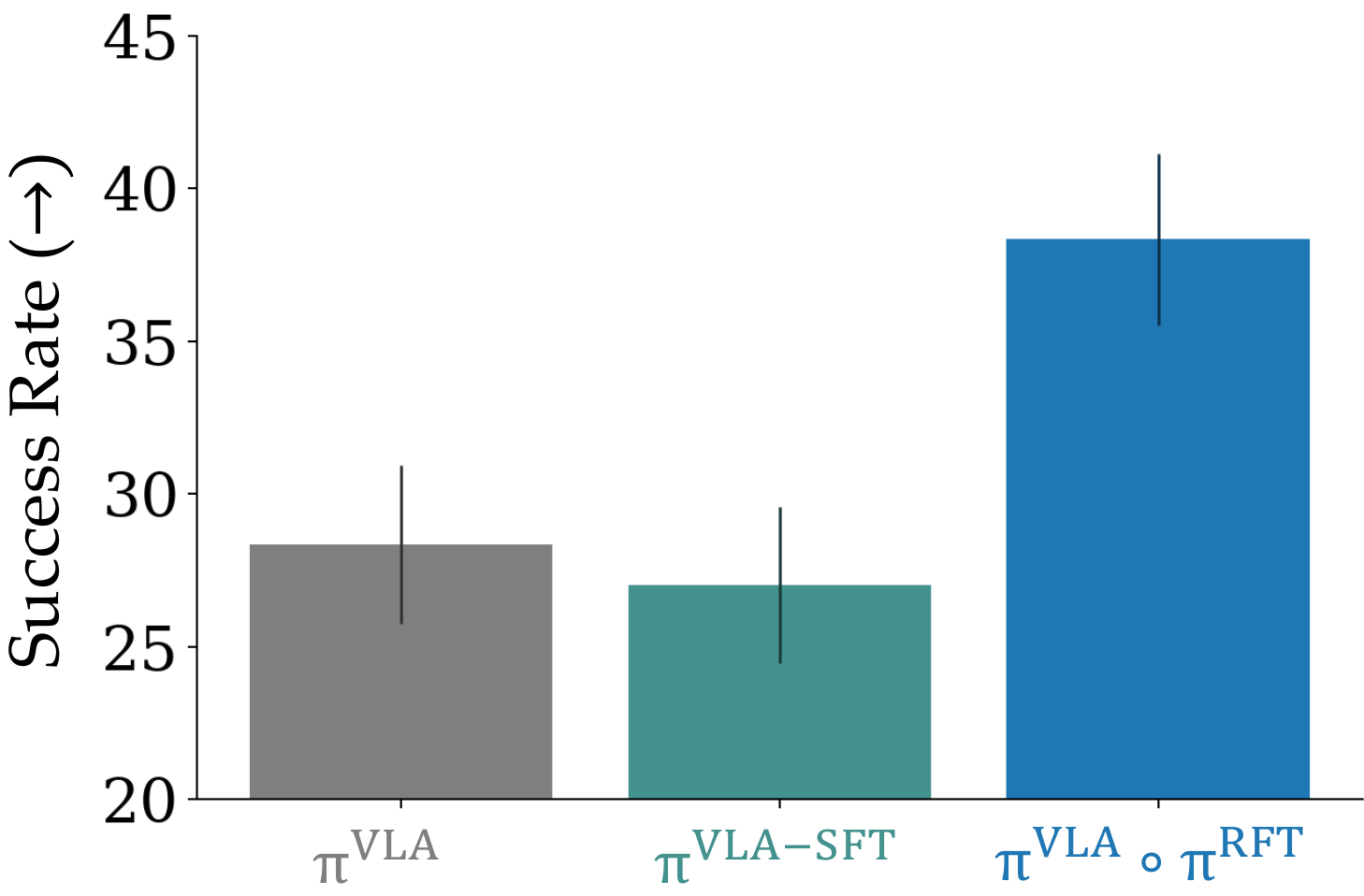}
    \caption{$\langpolicy$ steers $\vla$ in unseen behavior \textbf{Compose} tasks. Error bars are standard error over episodes across all perturbations.}
    \label{fig:novel}
    \vspace{-2em}
\end{wrapfigure}
\para{Evaluation and Metrics} We measure mean success rate (binary per rollout) with standard error across visual and semantic perturbations. We do 50 rollouts per task and perturbation combination (200 total combinations) in simulation episodes and 30 in hardware.

\noindent \textbf{Result: $\langpolicy$ Has Strong ID and OOD Performance.}
Figure~\ref{fig:baselines} shows how different iterations of the LFP perform in training and test environments. $\sftpolicy$ improves training success but degrades deployment success, indicating that narration alone is not enough to elicit the right behavior from the frozen VLA. Interactive search improves both training and deployment success, and conformalization further improves deployment success by refusing harmful steering, per our guarantee in Eq~\ref{eq:cp}.

\begin{wrapfigure}{l}{0.28\columnwidth}
    \vspace{-1.5em}
    \centering
    \includegraphics[width=\linewidth]{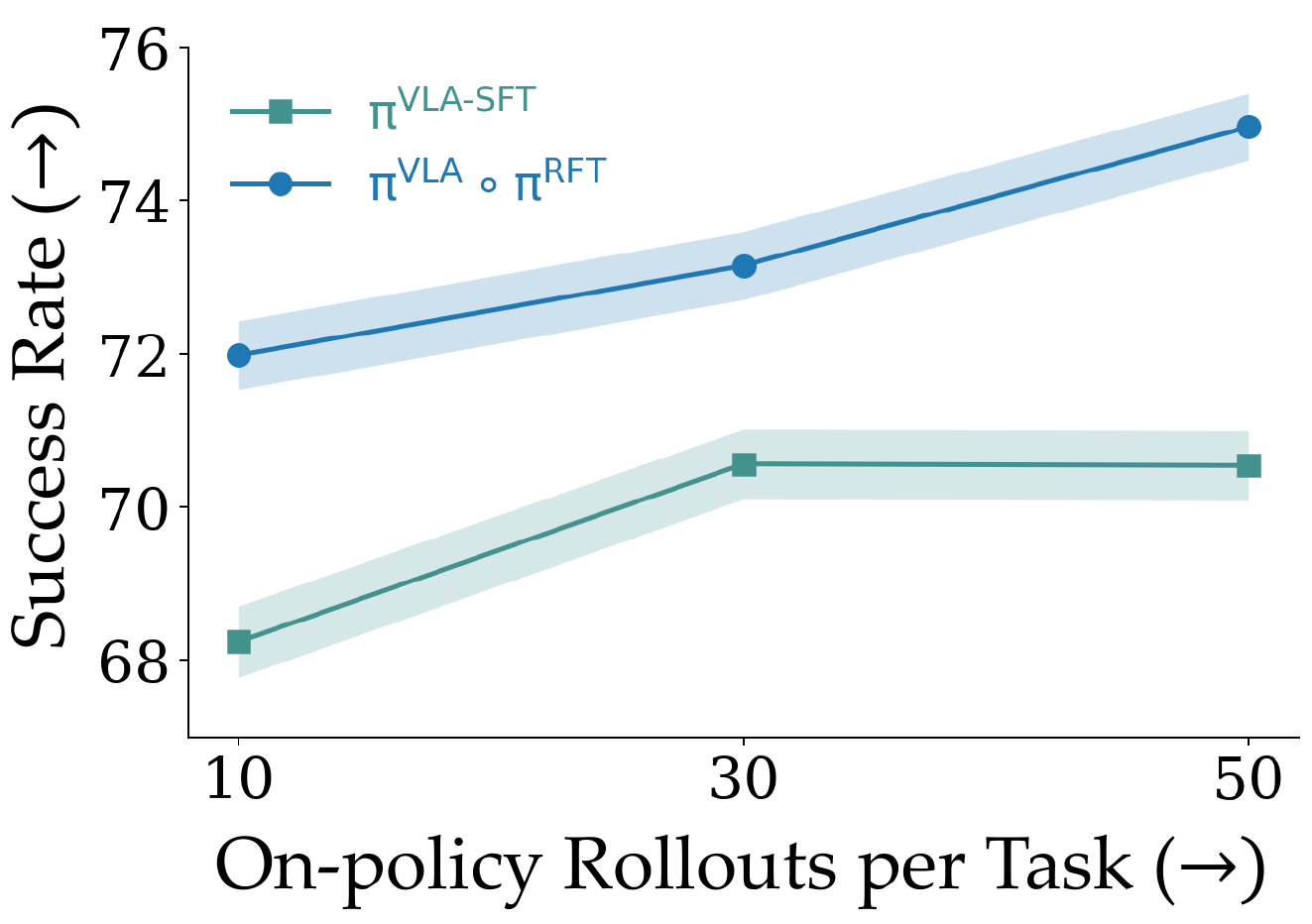}
    \caption{$\langpolicy$ learns more from same on-policy rollout data compared to $\vlasft$. Error bars are standard error over episodes across all perturbation combinations.}
    \label{fig:ftdata}
    \vspace{-1.5em}
\end{wrapfigure}
A comprehensive evaluation is reported in Table~\ref{tab:baselines-by-vood} in the Appendix. 
$\langpolicy$ and $\langpolicycal$ achieve the best success rate in all four simulation conditions, while $\vlasft$ improves on the training-like conditions but degrades on OOD scene conditions, supporting our hypothesis
that language steering generalizes better than directly adapting the action policy. Figure~\ref{fig:novel} shows that $\langpolicy$ also solves unseen behavior compositions, while direct VLA fine-tuning slightly degrades performance over $\vla$.
Hardware results in Fig.~\ref{fig:hardwareresults} show the same trend: $\langpolicy$ improves over $\vla$ on \textbf{CubeSort}, \textbf{CubeMug}, and \textbf{MarkerBlock}, while conformalization mitigates harmful steering on \textbf{Microwave}. On novel hardware tasks, $\langpolicy$ improves over both $\vla$ and $\sftpolicy$, suggesting that language steering transfers to novel tasks when the underlying VLA skills are present.

\noindent \textbf{Result: $\langpolicy$ Outperforms $\vlasft$ With Less On-Policy Rollout Data.} Figure~\ref{fig:ftdata} compares $\langpolicy$ and $\vlasft$ when trained with 10, 30, and 50 successful on-policy rollouts per task. This data scaling study shows a favorable exchange rate for language steering: $\langpolicy$ with only 10 rollouts already matches or exceeds $\vlasft$ trained with 50 rollouts, using \textit{\textbf{one-fifth}} as much fine-tuning data. Across all data scales, $\langpolicy$ consistently outperforms $\vlasft$ in nominal, visually perturbed, and semantically perturbed settings. Moreover, while $\vlasft$ plateaus by 50 successful on-policy rollouts, $\langpolicy$ continues to improve with additional data, suggesting that adapting the language interface uses limited fine-tuning data more efficiently than directly fine-tuning the VLA.

\subsection{Q2: When Should We (Not) Do Language Steering?}
We investigate when language steering should be applied, and whether refusal and conformalization can preserve its benefits by selectively intervening only when language edits are likely to help.

\noindent \textbf{Methods.}
We evaluate refusal and conformalization in both simulation and hardware. We compare three variants: no refusal $\langpolicynr$, which runs the same LFP as $\langpolicy$ but disables the refusal mechanism and always allows language steering; $\langpolicy$, which refuses using the uncalibrated threshold $\psi(\obs_0,\taskdesc)\geq 0$; and the calibrated variant $\langpolicycal$, which uses conformal prediction with $\alpha=0.10$.

\begin{table}[t]
    \centering
    \small
    \begin{tabular}{@{}llcccccc@{}}
    \toprule
    & & Accuracy ($\uparrow$) & TPR ($\uparrow$) & FPR ($\downarrow$) & TNR ($\uparrow$) & FNR ($\downarrow$) & Refusal Rate \\
    \midrule
    \multirow{2}{*}{Simulation} 
    & $\langpolicy$ 
    & $66.22\%$ 
    & $\mathbf{70.18\%}$ 
    & $38.92\%$ 
    & $61.08\%$ 
    & $\mathbf{29.82\%}$ 
    & $43.42\%$ \\

    & $\langpolicycal$ 
    & $\mathbf{77.74\%}$ 
    & $67.77\%$ 
    & $\mathbf{9.31\%}$ 
    & $\mathbf{90.69\%}$ 
    & $32.23\%$ 
    & $57.66\%$ \\
    \hline 
    \multirow{2}{*}{Hardware} 
    & $\langpolicy$ 
    & $84.72\%$ 
    & $\mathbf{100.00\%}$ 
    & $61.11\%$ 
    & $38.89\%$ 
    & $\mathbf{0.00\%}$ 
    & $9.72\%$ \\

    & $\langpolicycal$ 
    & $\mathbf{99.17\%}$ 
    & $99.63\%$ 
    & $\mathbf{2.22\%}$ 
    & $\mathbf{97.78\%}$ 
    & $0.37\%$ 
    & $24.72\%$ \\
    \bottomrule
    \end{tabular}
    \vspace{1em}
    \caption{\textbf{Confusion Matrix of $\langpolicy$'s Language Interventions.} After conformalization done at $\alpha=0.1$, FPR decreases from $38.92\%$ to $9.31\%$ in simulation and from $61.11\%$ to $2.22\%$ on hardware. This shows that conformalization controls FPR at or below the target level, mitigating harmful steering while only slightly decreasing TPR.}
    \label{tab:confusion}
    \vspace{-2em}
\end{table}
\para{Metrics}
In simulation, we sweep $\alpha\in\{0.05,0.10,\ldots,0.40\}$ and compare the target FPR to the empirical FPR on held-out test data of 10 episodes per condition, averaged over 5 randomized calibration/test splits. 
In both simulation and hardware, we report the confusion matrix of $\langpolicy$ and $\langpolicycal$ ($\alpha=0.10$): True Positive Rate (TPR; helpful task, steers), False Positive Rate (FPR; harmful task, steers), True Negative Rate (TNR; harmful task, refuses), and False Negative Rate (FNR; helpful task, refuses). We also report accuracy, refusal rate (the total percentage of episodes refused), and mean success rate across visual and semantic perturbations.

\begin{wrapfigure}{r}{0.6\columnwidth}
    \vspace{-0.4em}
    \centering
    \includegraphics[width=\linewidth]{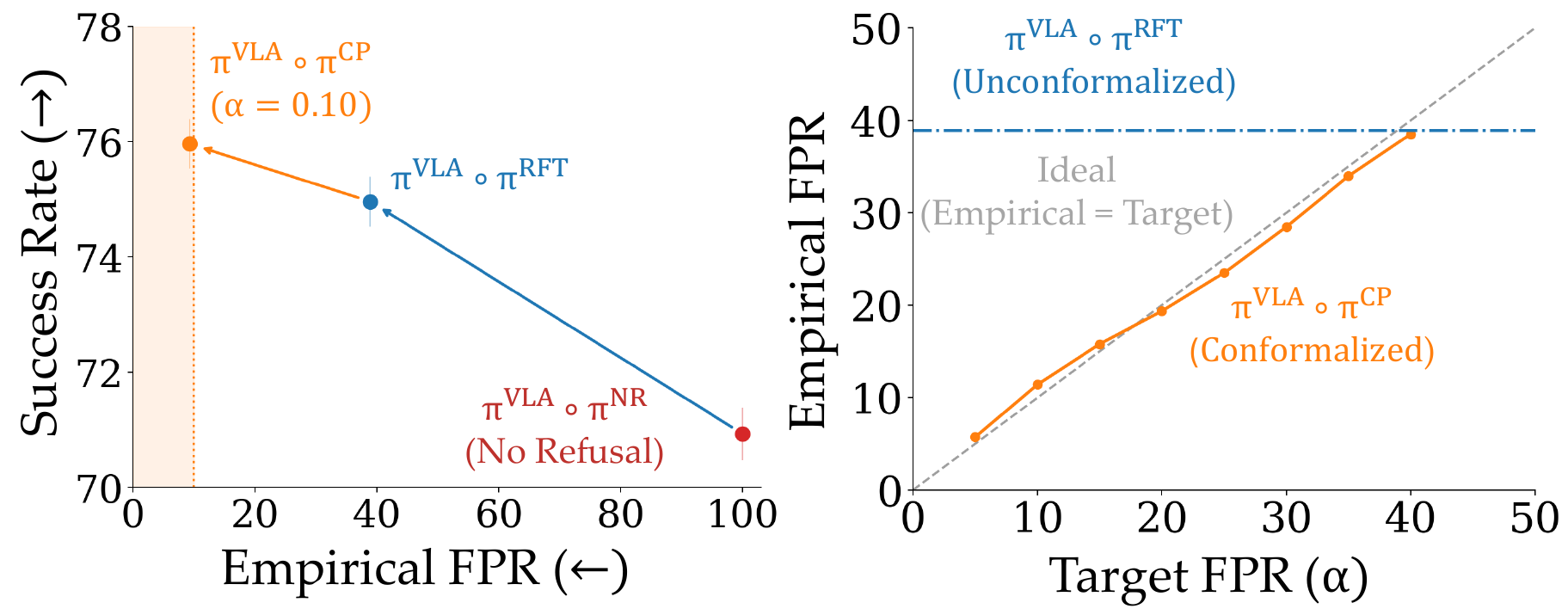}
    \caption{\textbf{Simulation Refusal Results}. \textit{Left}: success rate vs. FPR for no refusal $\langpolicynr$, $\langpolicy$, and $\langpolicycal$. Both refusal and conformalization help performance. \textit{Right}: reliability diagram shows that empirical FPR tracks the target $\alpha$ across 5 randomized splits.}
    \label{fig:refusal}
    \vspace{-2.0em}
\end{wrapfigure}
\noindent \textbf{Result: Refusal and Conformalization Improve Performance.}
Figure~\ref{fig:refusal} (left) shows that refusing language steering when it is harmful improves success rates, computed over 10,000 simulation rollouts pooled across the 200 visual and semantic perturbation combinations. \textit{Without} refusal, $\langpolicynr$ achieves $70.93\%$ success (red dot in Figure~\ref{fig:refusal}); adding refusal improves success to $74.96\%$, and conformalization further improves success to $75.96\%$ by reducing harmful steering.

\noindent \textbf{Result: Conformalization Controls Harmful Steering.}
Conformalization substantially reduces harmful steering. In simulation, FPR drops from $38.92\%$ to $9.31\%$ at $\alpha=0.1$, matching the target error rate; equivalently, TNR rises from $61.08\%$ to $90.69\%$. Hardware shows the same trend: FPR drops from $61.11\%$ to $2.22\%$, increasing accuracy from $84.72\%$ to $99.17\%$ with only a small TPR decrease ($100\%$ to $99.63\%$). Figure~\ref{fig:refusal}~(right) further shows that empirical FPR tracks the target $\alpha$ across conformalization levels. The full confusion matrix is shown in Table~\ref{tab:confusion}.

\begin{figure}
    \centering
    \includegraphics[width=1\linewidth]{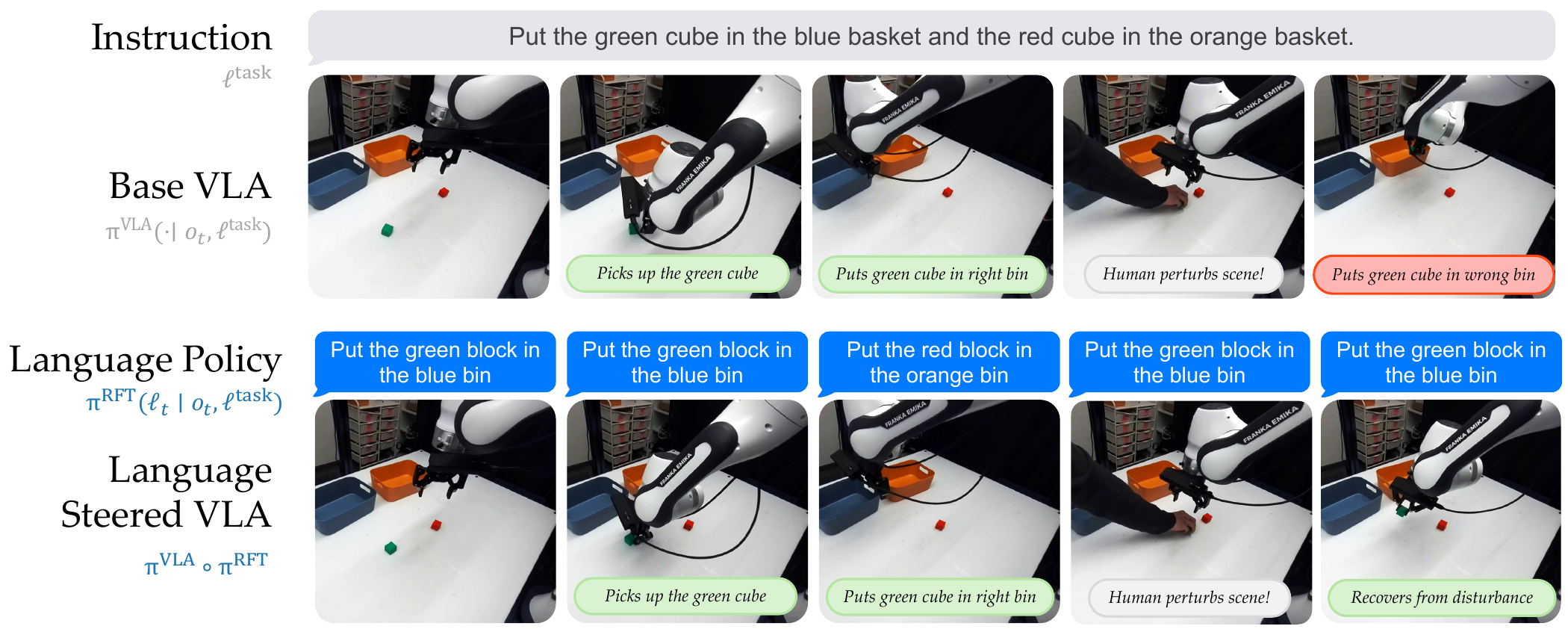}
    \caption{
    \textbf{Closed-loop Language Steering vs. Base VLA.} After the robot initially places the green cube correctly, a human moves the cube back into the scene. $\vla$ continues executing from the original task instruction and misplaces the cube, whereas $\langpolicy$ observes the changed state, outputs an updated language action $\lang_t$, and steers the frozen VLA to recover from the perturbation.
    }
    \label{fig:closedloop}
    \vspace{-1.7em}
\end{figure}
\subsection{Q3: How Does Open- vs. Closed-Loop Language Feedback Influence VLA Performance?}
We finally ablate two design choices that determine how language feedback is applied. First, we test whether closed-loop feedback is necessary by comparing the LFP against an open-loop static rewriter that edits the task instruction once and holds it fixed throughout the episode. Second, we study the structure of the local search space, comparing trajectory-level perturbations of complete language sequences against phrase-level perturbations of individual language actions.

\para{Methods}
In simulation, we compare our trajectory-level closed-loop search against two variants: open-loop search, which learns a \textit{single static} task description edit which is held constant throughout the episode, and a closed-loop \textit{phrase-level} search, which perturbs each $\lang_t$ independently.

\begin{wraptable}{r}{0.4\textwidth}
    \centering
    \small
    \vspace{-1em}
    \begin{tabular}{@{}l cc@{}}
        \toprule
        Method & SR (\%) & Refusal (\%) \\
        \midrule
        OL & $71.3 \pm 0.5$ & $43.5$ \\
        Phrase (CL) & $70.4 \pm 0.5 $ & $45.9$ \\
        Traj. (CL) & $\mathbf{75.0 \pm 0.4}$ & $\mathbf{43.4}$ \\
        \bottomrule
    \end{tabular}
    \caption{\textbf{Simulation Search Ablations.} OL: open-loop; CL: closed-loop. Mean success rate with standard error over episodes across all perturbation combinations are shown.}
    \label{tab:ablation_design}
    \vspace{-1em}
\end{wraptable}
\noindent \textbf{Result.}
We now demonstrate that closed loop steering provides a desirable form of robustness: the ability to recover from adversarial perturbations midway through the task. This behavior relies on both components of the LFP: closed-loop observation feedback and language-based steering for OOD generalization. Figure \ref{fig:closedloop} depicts this behavior. Midway through the task, a person removes the green block that the robot had previously placed in the blue bin. The base VLA picks up the green block, but places it in the wrong orange bin. Our closed-loop LFP updates its utterance based on the changed scene, generating the corrective instruction ``Put the green block in the blue bin'' and recovering from the perturbation by the action taken.

Table~\ref{tab:ablation_design} shows that trajectory-level closed-loop language feedback outperforms open-loop prompt search, indicating that language should adapt as the scene evolves. 
Trajectory-level search also outperforms phrase-level search, suggesting that sequence-level perturbations make the search space more tractable without sacrificing high quality candidates.

\subsection{Q4: Can We Learn Language Feedback Policies for Different Pre-Trained VLA's?}
\begin{wraptable}{r}{0.4\textwidth}
\vspace{-1.2em}
\centering
\small
\setlength{\tabcolsep}{6pt}
\renewcommand{\arraystretch}{0.9}
\begin{tabular}{l|cc}
\toprule
 & $\vla$ & $\langpolicy$ \\
\midrule
Sim. SR(\%) & $32.5\pm1.0$ & $\mathbf{40.6\pm1.0}$ \\
HW SR(\%)   & $40.0\pm8.9$ & $\mathbf{76.7\pm7.7}$ \\
\bottomrule
\end{tabular}
\caption{$\langpolicy$ deployed on MolmoAct2 in simulation (Visual-Viewpoint; 50 trials per task, 10 tasks) and hardware (CubeSort; 30 trials).}
\label{tab:molmoact2}
\vspace{-1.25em}
\end{wraptable}

To study the generality of our framework, we train and deploy $\langpolicy$ on an additional VLA beyond $\pi_{0.5}$: specifically, MolmoAct2 \cite{fang2026molmoact2actionreasoningmodels} in simulation (Visual-Viewpoint, 50 trials per task; 10 tasks total) and hardware (CubeSort, 30 trials). In simulation, we see a $+8.1$pp increase in success rate, and in hardware we see a $+36.7$pp increase, indicating that our training algorithm can ``learn what to say'' to multiple VLAs. 
% $\langpolicy$ is able to successfully steer in various $\vla$ models.
%\input{Sections/conclusion}
\section{Limitations}

%Although our conformalized language feedback policy improves base VLA performance and has strong harmlessness guarantees, limitations still remain. 
\smallskip 
\noindent \textbf{Complementary Steering Methods.} Steering via language is one of many ways to steer a pre-trained model. However, as we saw in this work, language is not always the best way to exert control authority over the VLA, which motivated our refusal mechanism. Future work should study the complementary role of editing language, editing the observation \cite{hancock2024runtimeobservationinterventionsmake}, steering the latents of the model \cite{haon2025mechanisticinterpretabilitysteeringvisionlanguageaction, wagenmaker2025steeringdiffusionpolicylatent}, or post-hoc editing the action generation directly \cite{nakamura2025train}.

\noindent \textbf{Language Search Strategy.} The language search space is still fundamentally a very large space to search over, and in this work we proposed a framework that searches locally around a fixed language sequence via paraphrasing. Future work should study how the language search space can be efficiently searched while still covering large, diverse, and realistic linguistic perturbations \cite{li2025elicitinglanguagemodelbehaviors}.

%\noindent \textbf{Inference Latency.} We incur additional inference latency at each replan step, which could limit dynamic task performance.

\noindent \textbf{Harmlessness Guarantees.} Harmlessness is guaranteed when perturbations follow the calibration distribution. Future work will explore how to stay robust beyond the calibration distribution.
%\clearpage
% The acknowledgments are automatically included only in the final and preprint versions of the paper.
\acknowledgments{We are grateful to Junwon Seo for his insights and discussions on conformal prediction and calibration. This research was supported in part by funding from the Defense Advanced Research Projects Agency (DARPA) SAFRON program Agreement No. HR0011-25-3-0204. GS was supported by an STTR grant. The views, opinions and/or findings expressed are those of the author and should not be interpreted as representing the official views or policies of DARPA or the U.S. Government.}

%===============================================================================

\bibliography{Bib/misc}  % .bib

\clearpage
\section{Appendix}\label{app:appendix}
\subsection{Related Work}
\noindent \textbf{Vision-Language-Action Models.} Vision-Language-Action (VLA) models combine language, vision, and robot control by mapping observations and task instructions to actions. Early VLAs predicted discrete action tokens \cite{kim2024openvlaopensourcevisionlanguageactionmodel}, while recent models use continuous or diffusion-style action heads conditioned on images and language \cite{intelligence2025pi05visionlanguageactionmodelopenworld,intelligence2026pi07steerablegeneralistrobotic}. Reasoning-based VLAs further introduce textual, visual, or latent intermediate representations to refine action generation \cite{zhao2025cotvlavisualchainofthoughtreasoning,lee2025molmoactactionreasoningmodels,fang2026molmoact2actionreasoningmodels,wu2026saysteeringvisionlanguageactionmodels}. Test-time steering methods guide VLA behavior through residual policies, external reward models, or verification signals \cite{wu2026saysteeringvisionlanguageactionmodels,jain2025smoothseaskilledsailor,xiao2025selfimprovingvisionlanguageactionmodelsdata}. Our contribution is complementary: we study how the frozen VLA can be steered via the language inputs alone, and learn a closed-loop language feedback policy with calibrated refusal.

\noindent \textbf{Steering Robots via Natural Language.} Natural-language robot control has long studied how robots ground instructions into symbols, plans, and executable behavior \cite{5453186,Tellex_Kollar_Dickerson_Walter_Banerjee_Teller_Roy_2011}. 
Recent work often adopts a hierarchical ``system 1/2" structure, where a high-level language or reasoning module interfaces with a low-level robot policy \cite{ren2023robotsaskhelpuncertainty,shi2024yellrobotimprovingonthefly,chen2026steerablevisionlanguageactionpoliciesembodied,intelligence2026pi07steerablegeneralistrobotic}. These methods typically rely on human feedback, demonstrations, or interventions to specify desired behavior. Prompt-optimization methods adapt language at test time \cite{myers2024policyadaptationlanguageoptimization,kwok2026scalingverificationeffectivescaling}, but do not distill the optimized language into a reusable closed-loop policy. Other recent work studies how to refine, preserve, or quantify language-following and task-switching in VLAs \cite{huang2026breakinglockinpreservingsteerability,chen2026resteerquantifyingrefiningsteerability}. We instead interactively search for language edits that elicit desired behavior from a frozen VLA, then distill them into a closed-loop language policy.

\noindent \textbf{VLA Robustness.} Reliable VLA deployment requires robustness to failures, distribution shift, and unsafe behavior. Prior work improves reliability through low-level action guardrails \cite{lin2025failsafereasoningrecoveryfailures,zhang2026safevlasafetyalignmentvisionlanguageaction}, latent activation steering \cite{mitra2025mechanisticfinetuningvisionlanguageactionmodels,haon2025mechanisticinterpretabilitysteeringvisionlanguageaction,buurmeijer2026observingcontrollingfeaturesvisionlanguageaction,swann2026sparseautoencodersrevealinterpretable}, or failure prediction probes \cite{gu2025safemultitaskfailuredetection}. Our work is complementary: we steer behavior through the language input, and use conformal prediction \cite{vovk2005algorithmic} to control the error rate of harmful steering under the calibration distribution.

\subsection{Method and Implementation Details}\label{app:implementation}
\begin{figure}[t]
    \centering
    \includegraphics[width=0.8\linewidth]{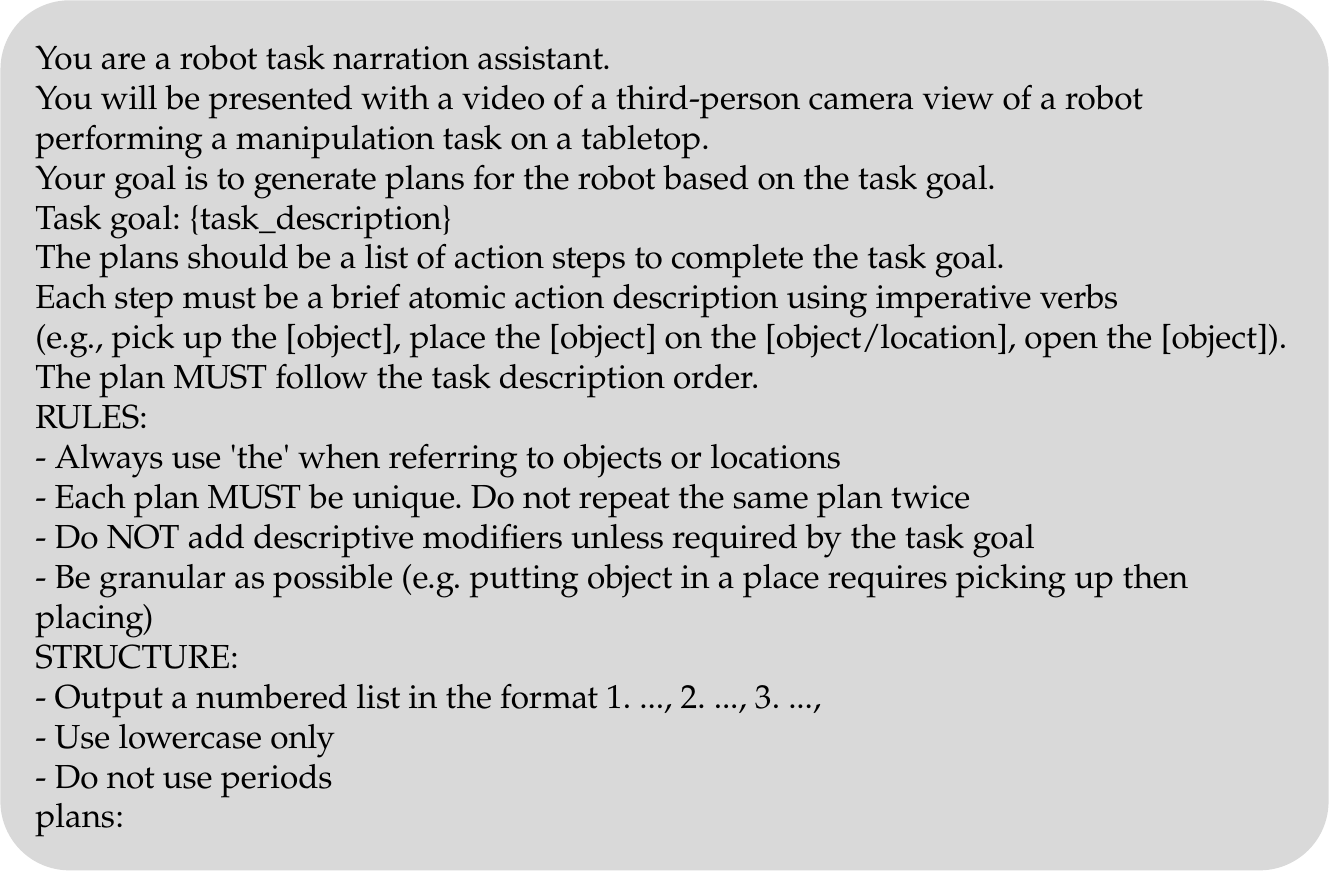}
    \caption{Prompt template for task decomposition.}
    \label{fig:decomposition}
\end{figure}

\begin{figure}[h]
    \centering
    \includegraphics[width=0.8\linewidth]{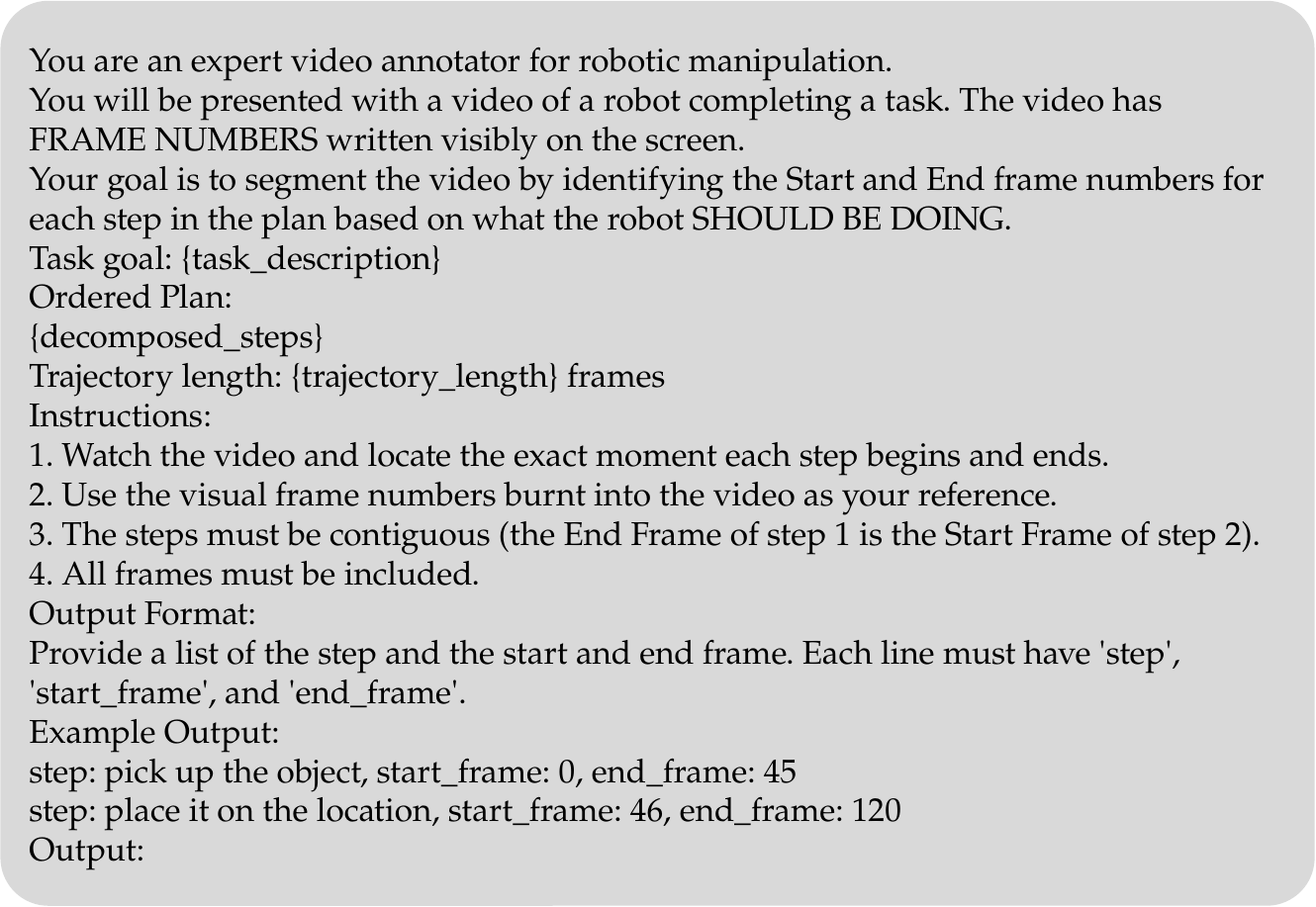}
    \caption{Prompt template for narration.}
    \label{fig:narration}
\end{figure}

\para{Narrated Fine-tuning}
To generate the narration prior for $\sftpolicyraw$, we leverage the video understanding capabilities of Molmo2-8B \cite{clark2026molmo2openweightsdata}. Given $\expdata=\{(\xi,\taskdesc)\}$, where each $\xi=(\obs_0,\ldots,\obs_T)$ is a visual trajectory paired with a task description, Molmo2-8B generates task decompositions of each video, as shown in figure \ref{fig:decomposition}. Given the most frequent decomposition, Molmo2-8B is then queried again to narrate the video. This ensures that each video gets a consistent temporal label for supervised fine-tuning. The prompt template for narration is shown in figure \ref{fig:narration}. Through decomposition and narration, $\nardata$ of tuples $(\obs,\taskdesc,\ell)$ is obtained. $\sftpolicyraw$ is fine-tuned via standard cross-entropy objective. Hyperparameters are detailed in table~\ref{tab:hp_sft}.
\vspace{-1em}

\begin{figure}[H]
    \centering
    \includegraphics[width=0.8\linewidth]{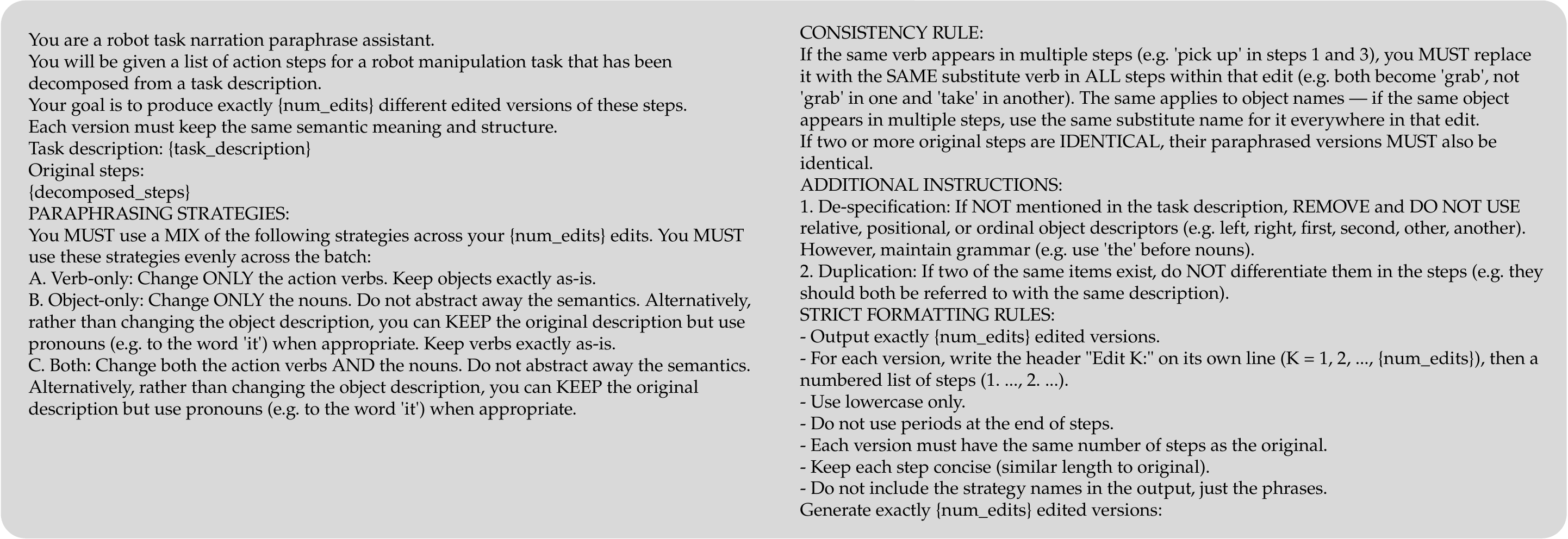}
    \caption{Prompt template for candidate sequence generation.}
    \label{fig:perturbation}
\end{figure}

\vspace{2em}
\para{Interactive Language Search}
Interactive language search generates local perturbations around $\sftpolicyraw$ and estimates the improvement of each perturbation. To generate local perturbations, we leverage the language capabilities of gpt-5.4 \cite{openai_gpt54thinking_2026} with high reasoning mode. Given the decomposition used to train $\sftpolicyraw$, we query gpt-5.4 with it along with the task description $\taskdesc$ to obtain $N=16$ trajectory-level semantic perturbations, where each perturbation rewrites the entire narrated sequence while preserving the intended task semantics. These perturbations paraphrase the verb, noun, and a mix of both. This forms an offline teacher-student pipeline: high-capacity models generate narrations and perturbations for search, which are then distilled into the lightweight closed-loop policy $\langpolicyraw$. The prompt template used for perturbation is depicted in figure \ref{fig:perturbation}. Interactive search selects the candidate sequence that maximizes expected improvement over the base VLA under the task distribution.

In practice, we estimate $\Delta(\taskdesc,c,\ell^{(n)}_{0:T})$ with Monte Carlo rollouts: steered rollouts are initialized from $p(\obs_0 \mid \taskdesc,\context)$ and executed using the frozen VLA with the SFT-produced language sequence perturbed to $\ell^{(n)}_{0:T}$, while base rollouts are initialized independently from the same distribution and executed using the original task instruction $\taskdesc$. In simulation, we first run $M=25$ rollouts per language sequence, keep the top $G=8$ candidates, and evaluate them with $M=75$ additional rollouts. In hardware, we run initially for $M=10$ rollouts, then $M=20$ additional rollouts for top $G=8$ candidates. For rejection fine-tuning, we use $K'=50$ successful on-policy demonstrations from the selected language sequence per task. Hyperparameters for expert iteration and improvement predictor training are denoted in table \ref{tab:hp_ei} and table \ref{tab:hp_value_head}, respectively. For all training runs, an Nvidia A6000 Ada was used.
\vspace{-0.5em}
\vspace{-0.5em}
\para{Conformalization}
Finally, we calibrate the intervention threshold on held-out visual and semantic perturbation episodes where the empirical task-level improvement is negative. In both simulation and hardware, we collect $N_{\mathrm{cal}}=20$ calibration samples for each perturbation combination.

\begin{figure}[H]
    \centering
    \includegraphics[width=0.8\linewidth]{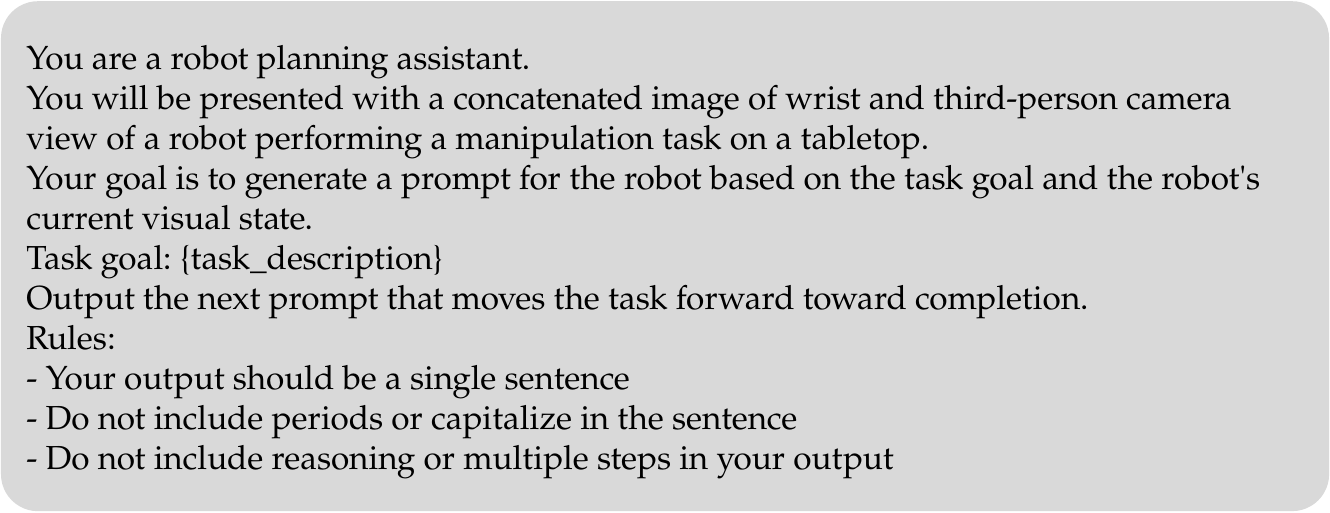}
    \caption{Prompt template for closed-loop steering.}
    \label{fig:steer}
\end{figure}

\para{Deployment-time Steering}
At deployment time, we query $\langpolicyraw$ at every replan step of the VLA. In LIBERO, this is every 5 environment steps, and in DROID, this is every 8 environment steps. The prompt template for closed-loop steering is shown in Fig.~\ref{fig:steer}.

\subsection{Conformal Guarantee for Calibrated Refusal}
\label{app:conformal_proof}

We prove the class-conditional guarantee in Eq.~\ref{eq:cp}. Recall that the calibration set is
$\caldata=\{(X^i,c^i,\widehat{\Delta}^i)\}_{i=1}^{N_{\mathrm{cal}}}$, where
$X^i=(\obs^i_0,\taskdesc[,i])$ and $\widehat{\Delta}^i$ is the empirical language improvement of steering over the base VLA. We define
$Y^i=\mathbb{I}\{\widehat{\Delta}^i\geq 0\}$, so $Y^i=0$ denotes a harmful steering example, and let
$s^i=\psi(X^i)$ be the learned improvement score.

Let $\mathcal{I}^0=\{i:Y^i=0\}$ denote the harmful calibration examples, and let
$m=|\mathcal{I}^0|$. Let
$s^0_{(1)}\leq s^0_{(2)}\leq \cdots \leq s^0_{(m)}$
denote the sorted scores $\{s^i:i\in\mathcal{I}^0\}$. We set the calibrated steering threshold to the finite-sample conformal quantile
\begin{equation}
\label{eq:conformal_quantile}
    \calibval
    =
    s^0_{(k)},
    \qquad
    k=\left\lceil (m+1)(1-\alpha)\right\rceil,
\end{equation}
with the convention that $\calibval=+\infty$ if $k>m$.

Consider a deployment example $(X,Y)$ drawn exchangeably with the calibration examples, and condition on the event $Y=0$. Let $s=\psi(X)$ denote its improvement score. Under the standard conformal exchangeability assumption, the scores
\begin{equation*}
    \{s^i:i\in\mathcal{I}^0\}\cup\{s\}
\end{equation*}
are exchangeable conditional on $Y=0$. Therefore, the rank of $s$ among these $m+1$ harmful-class scores is uniformly distributed up to ties. The calibrated rule incorrectly allows steering on a harmful example only when
\begin{equation*}
    s=\psi(X)\geq \calibval.
\end{equation*}
By the definition of the conformal quantile in Eq.~\ref{eq:conformal_quantile}, this event occurs only when the test score lies above the $(1-\alpha)$ quantile of the harmful-class score distribution. Hence,
\begin{equation}
    \mathbb{P}\left(
    \psi(X)\geq \calibval
    \mid Y=0
    \right)
    \leq \alpha.
\end{equation}
Thus, among examples where steering is empirically harmful, the probability that the calibrated rule incorrectly permits steering is at most $\alpha$.

\begin{table}[H]
  \centering
  \small
  \begin{tabular}{@{}ll@{}}
    \toprule
    \textbf{Category} & \textbf{Value} \\
    \midrule
    \multicolumn{2}{l}{\textit{Model}} \\
    Base model & Qwen3-VL-4B-Instruct \\
    Adapter & LoRA (r=64, $\alpha=128$, dropout=0) \\
    Target modules & q, k, v, o, gate, up, down\_proj \\
    \midrule
    \multicolumn{2}{l}{\textit{Data}} \\
    Inputs & Wrist + agent view images \\
    Supervision & Per-step $(o_t, \ell_t)$ demonstrations \\
    Loss & Cross-entropy \\
    \midrule
    \multicolumn{2}{l}{\textit{Optimization}} \\
    Optimizer & AdamW \\
    Learning rate & $1\times10^{-5}$ \\
    Weight decay & $0.01$ \\
    Batch size & 1 (no accumulation) \\
    Epochs & 1 \\
    Precision & bfloat16 \\
    \bottomrule
  \end{tabular}
  \caption{Narrated fine-tuning hyperparameters.}
  \label{tab:hp_sft}
\end{table}

\begin{table}[H]
  \centering
  \nopagebreak
  \begin{minipage}[t]{0.48\linewidth}
    \centering
    \small
    \begin{tabular}{@{}ll@{}}
      \toprule
      \textbf{Category} & \textbf{Value} \\
      \midrule
      \multicolumn{2}{l}{\textit{Model}} \\
      Base model & Qwen3-VL-4B-Instruct \\
      Initialization & SFT LoRA checkpoint (frozen base) \\
      \midrule
      \multicolumn{2}{l}{\textit{Data}} \\
      Inputs & Wrist + agent view images \\
      Supervision & Per-step $(o_t, \ell_t)$ demonstrations \\
      Loss & Cross-entropy \\
      \midrule
      \multicolumn{2}{l}{\textit{Optimization}} \\
      Optimizer & AdamW \\
      Learning rate & $1\times10^{-5}$ \\
      Weight decay & 0 \\
      Batch size & 1 \\
      Epochs & 1 \\
      Precision & bfloat16 \\
      \bottomrule
    \end{tabular}
    \caption{Expert Iteration hyperparameters.}
    \label{tab:hp_ei}
  \end{minipage}
  \hfill
  \begin{minipage}[t]{0.48\linewidth}
    \centering
    \small
    \begin{tabular}{@{}ll@{}}
      \toprule
      \textbf{Category} & \textbf{Value} \\
      \midrule
      \multicolumn{2}{l}{\textit{Model}} \\
      Backbone & Frozen $\langpolicy$ \\
      Feature & Final hidden state \\
      Head & MLP ($d \rightarrow 64 \rightarrow 1$, ReLU, dropout=0.1) \\
      Feature norm & Per-dimension z-score \\
      \midrule
      \multicolumn{2}{l}{\textit{Data}} \\
      Inputs & Hidden state features \\
      Supervision & Task improvement \\
      Target & $\hat{\Delta}_i$ \\
      \midrule
      \multicolumn{2}{l}{\textit{Optimization}} \\
      Optimizer & Adam \\
      Learning rate & $1\times10^{-3}$ \\
      Weight decay & $1\times10^{-3}$ \\
      Loss & MSE \\
      Epochs & up to 200 (early stopping) \\
      Validation split & 20\% \\
      \bottomrule
    \end{tabular}
    \caption{improvement predictor hyperparameters.}\label{tab:hp_value_head}
  \end{minipage}
\end{table}

\begin{figure}
    \centering
    \includegraphics[width=1\linewidth]{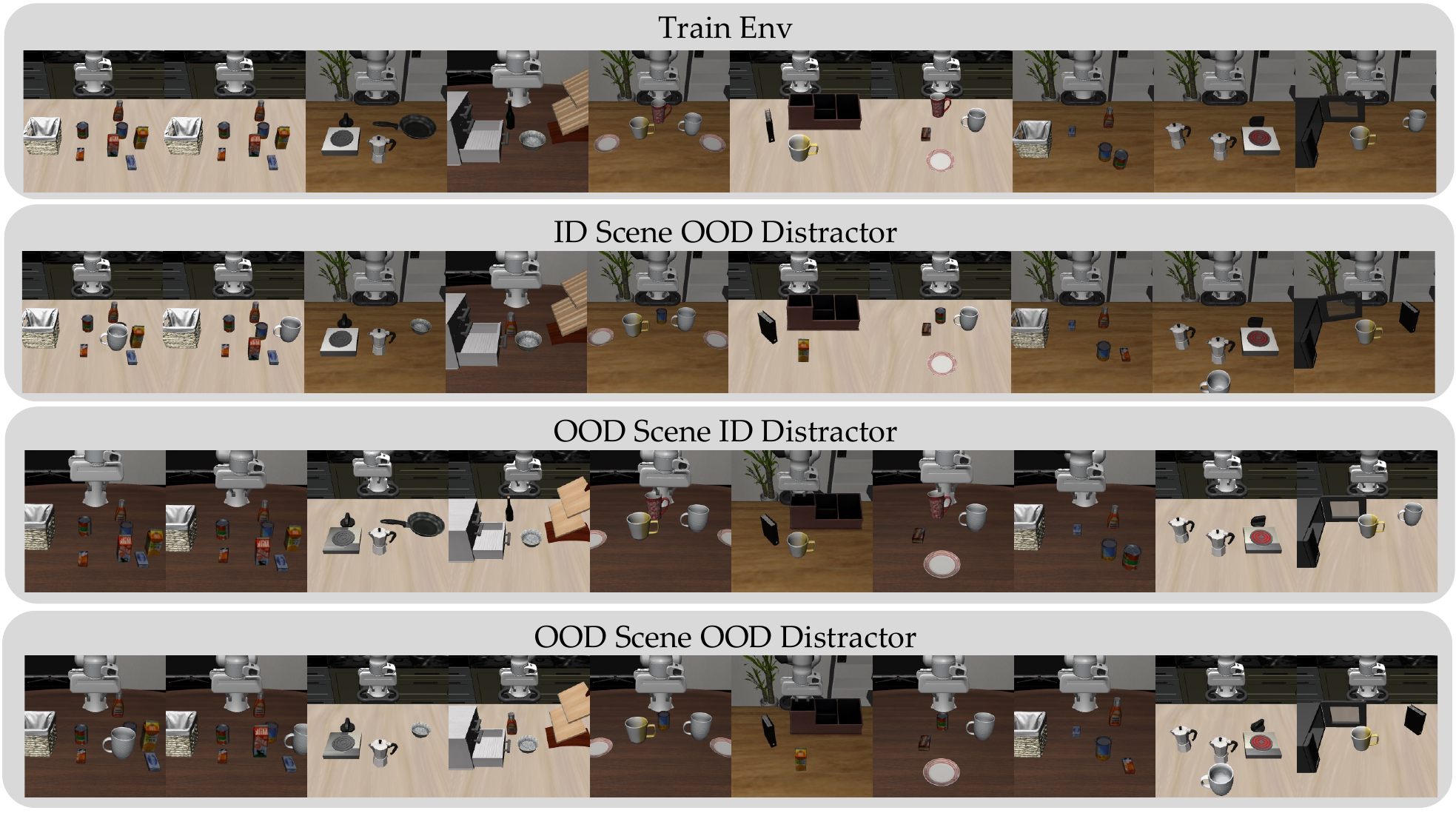}
    \caption{10 LIBERO-OOD tasks (left to right), as well as their perturbations (top to bottom).}
    \label{fig:libero}
\end{figure}

\begin{table*}[h!]
\centering
\small
\renewcommand{\arraystretch}{1.2} % Gives rows a little more vertical breathing room
\setlength{\tabcolsep}{8pt}

\begin{tabularx}{\textwidth}{c >{\raggedright\arraybackslash}X}
\toprule
\textbf{Task} & \textbf{Nominal Task Instruction} \\
\midrule
0 & put both the alphabet soup and the tomato sauce in the basket \\
1 & put both the cream cheese box and the butter in the basket \\
2 & turn on the stove and put the moka pot on it \\
3 & put the black bowl in the bottom drawer of the cabinet and close it \\
4 & put the white mug on the left plate and put the yellow and white mug on the right plate \\
5 & pick up the book and place it in the back compartment of the caddy \\
6 & put the white mug on the plate and put the chocolate pudding to the right of the plate \\
7 & put both the alphabet soup and the cream cheese box in the basket \\
8 & put both moka pots on the stove \\
9 & put the yellow and white mug in the microwave and close it \\
\bottomrule
\end{tabularx}
\caption{LIBERO-10 task instructions.}
\label{tab:libero10-nominal-instructions}
\end{table*}

\begin{table*}[h!]
\centering
\small
\renewcommand{\arraystretch}{1.15} 
\setlength{\tabcolsep}{6pt}       

\begin{tabularx}{\textwidth}{c >{\raggedright\arraybackslash}X}
\toprule
\textbf{Task} & \textbf{Perturbed Instructions} \\
\midrule

0 & \textbf{SOOD 1:} place the alphabet soup in the basket and the tomato sauce as well \\
  & \textbf{SOOD 2:} put both the can of soup and can of sauce in the basket \\
  & \textbf{SOOD 3:} put away the alphabet soup and tomato sauce in the basket \\
  & \textbf{SOOD 4:} place the can of soup in the basket and the can of sauce in the basket \\
\midrule

1 & \textbf{SOOD 1:} pick up both the cream cheese box and the butter and place them in the basket \\
  & \textbf{SOOD 2:} put both the box of cheese and the box of butter in the basket \\
  & \textbf{SOOD 3:} put away the cheese box and butter box in the basket \\
  & \textbf{SOOD 4:} stash the cream cheese and butter \\
\midrule

2 & \textbf{SOOD 1:} switch the stove on and set the moka pot on the stove \\
  & \textbf{SOOD 2:} turn on the cooktop and place the moka machine on it \\
  & \textbf{SOOD 3:} put the moka pot over the heat \\
  & \textbf{SOOD 4:} heat up the stove and put the moka pot on it \\
\midrule

3 & \textbf{SOOD 1:} move the black bowl to the bottom drawer and close the drawer of the cabinet \\
  & \textbf{SOOD 2:} put the middle bowl to the lowest drawer and close it \\
  & \textbf{SOOD 3:} put away the black bowl in the bottom drawer \\
  & \textbf{SOOD 4:} stash away the middle bowl in the lower cabinet \\
\midrule

4 & \textbf{SOOD 1:} place two mugs on the plates, left one is the white mug and the right one is the yellow and white mug \\
  & \textbf{SOOD 2:} put the pure white cup on the left plate and put the other one with the yellow handle on the right plate \\
  & \textbf{SOOD 3:} serve the cups \\
  & \textbf{SOOD 4:} serve the mugs \\
\midrule

5 & \textbf{SOOD 1:} grab the standing book and transfer it to the back compartment of the caddy \\
  & \textbf{SOOD 2:} pick up the right book and put it in the rear part of the caddy \\
  & \textbf{SOOD 3:} place the book in the back of the caddy \\
  & \textbf{SOOD 4:} put away the book in the back part \\
\midrule

6 & \textbf{SOOD 1:} set the white mug on the plate with chocolate pudding to be placed to the right \\
  & \textbf{SOOD 2:} put the pure white cup on the middle plate and put the brown chocolate to the right of the plate \\
  & \textbf{SOOD 3:} place the white cup on the plate and the chocolate to the right \\
  & \textbf{SOOD 4:} set the white mug on the plate and the chocolate pudding set on the right \\
\midrule

7 & \textbf{SOOD 1:} put both the can of soup and the box of cheese in the basket \\
  & \textbf{SOOD 2:} move the two objects, alphabet soup and cream cheese box, to the basket \\
  & \textbf{SOOD 3:} add the cream cheese box and the alphabet soup to the collection \\
  & \textbf{SOOD 4:} grab the alphabet soup and the cream cheese box and throw them in the basket \\
\midrule

8 & \textbf{SOOD 1:} transfer both moka pots from the table to the stove \\
  & \textbf{SOOD 2:} put both the moka coffee makers on the cooktop \\
  & \textbf{SOOD 3:} put the coffee makers where they can heat up \\
  & \textbf{SOOD 4:} heat up both moka pots \\
\midrule

9 & \textbf{SOOD 1:} place the yellow and white mug inside the microwave, then shut the door \\
  & \textbf{SOOD 2:} put the middle mug inside the microwave and close the door \\
  & \textbf{SOOD 3:} heat up the yellow and white mug by placing it and shutting the door \\
  & \textbf{SOOD 4:} heat up the yellow and white mug in the microwave \\
\bottomrule
\end{tabularx}
\caption{LIBERO-OOD semantic perturbations (SOOD1-2) and newly added semantic perturbations (SOOD 3-4).}
\label{tab:libero10-perturbed-instructions}
\end{table*}

\subsection{Benchmark Details}\label{app:benchmark}
\para{LIBERO-OOD}
LIBERO-OOD \cite{wu2026saysteeringvisionlanguageactionmodels} contains 2 visually perturbed environments, \textbf{Visual-Scene} (OOD Scene, OOD Distractor), which adds novel distractor objects, and \textbf{Visual-Viewpoint} (Train Env), which changes the background and camera viewpoints. It also contains 2 semantic perturbations over the original task description $\taskdesc$. We further add \textbf{Visual-Scene-Viewpoint} (ID Scene, OOD Distractor), which combines novel distractors with Visual-Viewpoint, as well as 2 additional semantic perturbations per task. The full suite of tasks are visualized in figure \ref{fig:libero} (tasks 0-9 are shown left to right), and their respective nominal task instructions are shown in table \ref{tab:libero10-nominal-instructions}. LIBERO-10 is denoted as OOD Scene ID Distractor. The perturbed instructions are shown in table \ref{tab:libero10-perturbed-instructions}. Novel behavior composition tasks are shown in figure \ref{fig:novel}, and their corresponding task instructions are shown in table \ref{tab:bood-instructions}.

\begin{figure}
    \centering
    \includegraphics[width=1\linewidth]{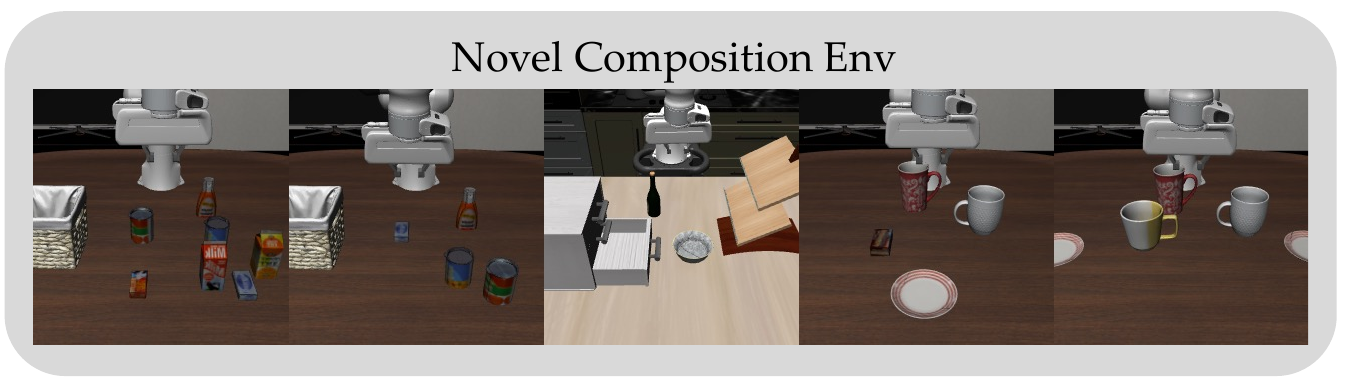}
    \caption{\textbf{LIBERO-OOD Novel Behavior Composition tasks.} Tasks 10-16 correspond to the 1st environment, tasks 17-18 to the 2nd, task 19 to the 3rd, task 20-21 to the 4th, and task 22 to the 5th.}
    \label{fig:bood}
\end{figure}

\begin{table}[h!]
\centering
\small
\renewcommand{\arraystretch}{1.05}
\setlength{\tabcolsep}{3pt}
\begin{tabularx}{\columnwidth}{c X}
\toprule
\textbf{Task} & \textbf{Instruction} \\
\midrule
10 & put both the cream cheese box and the tomato sauce in the basket \\
11 & put both the milk and the tomato sauce in the basket \\
12 & put both the orange juice and the tomato sauce in the basket \\
13 & put both the alphabet soup and the butter in the basket \\
14 & put both the orange juice and the butter in the basket \\
15 & put both the tomato sauce and the butter in the basket \\
16 & put both the milk and the butter in the basket \\
17 & put both the ketchup and the cream cheese box in the basket \\
18 & put both the tomato sauce and the cream cheese box in the basket \\
19 & put the wine bottle in the bottom drawer of the cabinet and close it \\
20 & put the red mug on the plate and put the chocolate pudding to the right of the plate \\
21 & put the red mug on the plate and put the chocolate pudding to the left of the plate \\
22 & put the red mug on the left plate and put the yellow and white mug on the right plate \\
\bottomrule
\end{tabularx}
\caption{BOOD (behavior composition) task instructions for LIBERO-10 tasks.}
\label{tab:bood-instructions}
\end{table}

\begin{figure}[h!]
    \centering
    \includegraphics[width=1\linewidth]{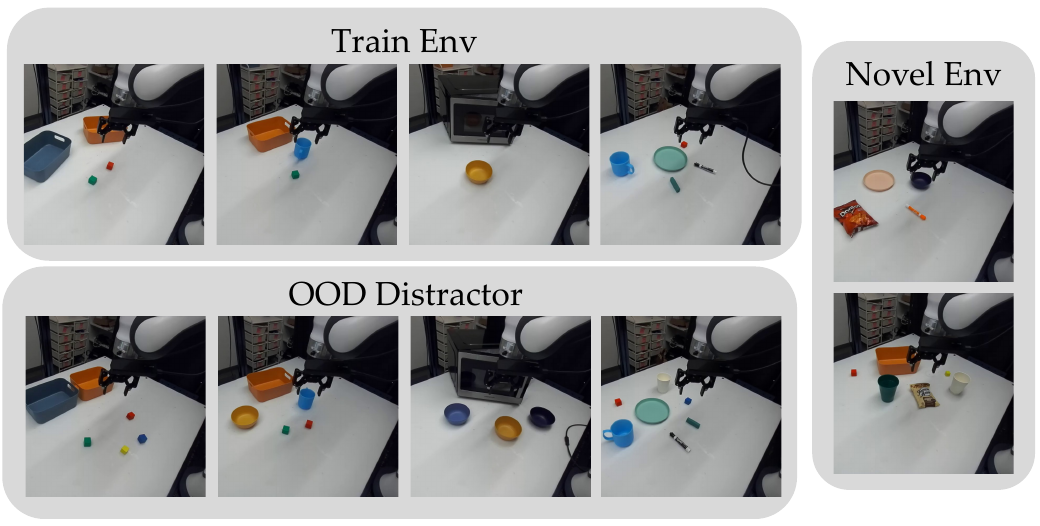}
    \caption{\textbf{Hardware Environments.} Left: training environments and their visual perturbations. \textbf{CubeSort, CubeMug, Microwave,} and \textbf{MarkerBlock} shown left to right. Right: novel environments \textbf{MarkerBowl} (top) and \textbf{ChipsCup} (bottom).}
    \label{fig:hardwareenvs}
\end{figure}

\begin{table}[h!]
  \centering
  \small
  \begin{tabular}{@{}lp{0.38\linewidth}p{0.38\linewidth}@{}}
    \toprule
    Task & Nominal & Semantic Perturbation \\
    \midrule
    \textbf{CubeSort} & put the green cube in the blue basket and the red cube in the orange basket & put away the green cube and the red cube in similar color baskets \\
    \textbf{CubeMug} & place the green cube in the blue mug then the blue mug in the orange basket & put away the cube in the mug and the mug in the bin \\
    \textbf{Microwave} & place the orange bowl on top of the microwave and close the microwave door & put the bowl on top of the oven and close the door \\
    \textbf{MarkerBlock} & place the marker and green block on the plate & put away the pen and green block in the bowl \\
    \textbf{MarkerBowl} & put the marker in the blue bowl and put the blue bowl on the plate & --- \\
    \textbf{ChipsCup} & put the chips and green cup in the basket & --- \\
    \bottomrule
  \end{tabular}
  \caption{Hardware task descriptions (nominal and semantic perturbation).}
  \label{tab:droid_task_descriptions}
\end{table}

\para{Hardware}
We evaluate our framework on 4 distinctive training environments, along with a visual and semantic perturbation per environment. The visual perturbation adds new distractor objects to the scene. We additionally evaluate on 2 novel tasks that are unseen during training. \textbf{CubeSort} requires the robot to place a green cube in the blue basket and a red cube in the orange basket, \textbf{CubeMug} requires the robot to place a green cube in the blue mug, then the blue mug in an orange basket, \textbf{Microwave} requires the robot to place the orange bowl on top of the microwave and close the microwave door, \textbf{MarkerBlock} requires the robot to place a marker and a green block on a plate, \textbf{MarkerBowl} requires the robot to put the marker in the blue bowl and the bowl on a plate, and \textbf{ChipsCup} requires the robot to put the chips and green cup in the basket. Hardware tasks are shown in figure \ref{fig:hardwareenvs}, and their respective task descriptions (along with semantic perturbations) are shown in table \ref{tab:droid_task_descriptions}.

\para{Result Breakdown}
A detailed breakdown per visual perturbation of the results from figure \ref{fig:baselines} are shown in table \ref{tab:baselines-by-vood}. In Visual-Viewpoint (train env.) and Visual-Scene-Viewpoint (ID Scene, OOD Distractor), $\langpolicy$ is able to improve performance the most over $\vla$ by $12.7$ pp and $8.3$ pp respectively. In LIBERO-10, scenarios where $\vla$ is already performs well (since it was fine-tuned on LIBERO-10 data), $\langpolicy$ and $\langpolicycal$ are able to maintain and slightly increase performance, whereas $\vlasft$ hurts overall performance. In Visual-Scene, although $\langpolicy$ hurts performance relative to $\vla$ by $1.3$ pp, conformalization is able to mitigate harmful interventions and helps overall performance by $1.1$ pp. Although $\vlasft$ is able to slightly increase performance in Visual-Viewpoint and Visual-Scene-Viewpoint, it overfits, hurting performance in scenarios it was previously performant in (LIBERO-10 and Visual-Scene). A breakdown of the novel behavior composition results (per task) are shown in table \ref{tab:bood-baselines}. $\langpolicy$ is able to successfully steer in most tasks, particularly on task 13. However, it is not able to steer on task 15, while $\vla$ and $\vlasft$ are able to perform better. Note that figure \ref{fig:bood} includes the first 6 tasks from the novel task composition suite, as tasks 16-22 are zero for all baselines. 

Hardware results are broken down by task and perturbation condition in Table~\ref{tab:hw-sr-by-task}. $\langpolicy$ substantially improves over the base VLA on the steerable training tasks \textbf{CubeSort}, \textbf{CubeMug}, and \textbf{MarkerBlock}, and also transfers to the novel tasks \textbf{MarkerBowl} and \textbf{ChipsCup}. In contrast, \textbf{Microwave} is a less steerable task: uncalibrated steering hurts performance under semantic perturbation, while $\langpolicycal$ refuses harmful interventions and recovers performance above the base VLA.

\begin{table}[h!]
\centering
\small
\setlength{\tabcolsep}{4pt}
\renewcommand{\arraystretch}{1.15}
\begin{tabular}{lcccccc}
\toprule
\textbf{Condition} & $\vla$ & $\vlasft$ & $\basevlm$ & $\sftpolicy$ & $\langpolicy$ & $\langpolicycal$ \\
\midrule
\textbf{Train Env} & 51.2 & 60.6 & 53.1 & 55.0 & \textbf{63.9} & 63.8 \\
\textbf{ID Scene, OOD Distractor} & 46.7 & 50.7 & 45.7 & 47.3 & \textbf{55.0} & 54.2 \\
\textbf{OOD Scene, ID Distractor} & 92.4 & 88.2 & 92.4 & 89.1 & 92.5 & \textbf{95.0} \\
\textbf{OOD Scene, OOD Distractor} & 89.8 & 82.6 & 89.5 & 81.2 & 88.5 & \textbf{90.9} \\
\bottomrule
\end{tabular}
\caption{Baseline success rates (\%) by visual perturbation condition, pooled over tasks and semantic perturbations.}
\label{tab:baselines-by-vood}
\end{table}

\begin{table}[h!]
\centering
\scriptsize
\setlength{\tabcolsep}{3.5pt}
\renewcommand{\arraystretch}{1.15}
\begin{tabular}{lccccccccccccc}
\toprule
\textbf{Method} & 10 & 11 & 12 & 13 & 14 & 15 & 16 & 17 & 18 & 19 & 20 & 21 & 22 \\
\midrule
$\vla$ & 88.0 & 0.0 & 2.0 & 36.0 & \textbf{4.0} & 40.0 & 0.0 & 0.0 & 0.0 & 0.0 & 0.0 & 0.0 & 0.0 \\
$\vlasft$ & 94.0 & 2.0 & \textbf{10.0} & 10.0 & 0.0 & \textbf{46.0} & 0.0 & 0.0 & 0.0 & 0.0 & 0.0 & 0.0 & 0.0 \\
$\langpolicy$ & \textbf{98.0} & \textbf{4.0} & \textbf{10.0} & \textbf{88.0} & 0.0 & 30.0 & 0.0 & 0.0 & 0.0 & 0.0 & 0.0 & 0.0 & 0.0 \\
\bottomrule
\end{tabular}
\caption{novel task composition per-task success rates (\%). Tasks 16--22 have 0\% SR (no method succeeds).}
\label{tab:bood-baselines}
\end{table}

\begin{table}[h!]
\centering
\small
\setlength{\tabcolsep}{4pt}
\renewcommand{\arraystretch}{1.15}
\begin{tabular}{llcccc}
\toprule
\textbf{Task} & \textbf{Condition} & $\vla$ & $\sftpolicy$ & $\langpolicy$ & $\langpolicycal$ \\
\midrule
\multirow{3}{*}{CubeSort}
  & Nominal  & 56.7 & 80.0 & \textbf{93.3} & 90.0 \\
  & VOOD     & 13.3 & 43.3 & \textbf{70.0} & \textbf{70.0} \\
  & SOOD     &  6.7 & 26.7 & 60.0 & \textbf{63.3} \\
\midrule
\multirow{3}{*}{CubeMug}
  & Nominal  & 36.7 & 63.3 & \textbf{83.3} & 76.7 \\
  & VOOD     & 30.0 & 40.0 & 66.7 & \textbf{73.3} \\
  & SOOD     & 16.7 & 36.7 & \textbf{70.0} & \textbf{70.0} \\
\midrule
\multirow{3}{*}{Microwave}
  & Nominal  & 73.3 & 66.7 & \textbf{76.7} & 73.3 \\
  & VOOD     & 46.7 & 16.7 & \textbf{50.0} & \textbf{50.0} \\
  & SOOD     & 66.7 & 40.0 & 46.7 & \textbf{70.0} \\
\midrule
\multirow{3}{*}{MarkerBlock}
  & Nominal  & 33.3 & 80.0 & \textbf{90.0} & \textbf{90.0} \\
  & VOOD     & 36.7 & 60.0 & \textbf{76.7} & \textbf{76.7} \\
  & SOOD     & 30.0 & 43.3 & 63.3 & \textbf{70.0} \\
\midrule
MarkerBowl & Nominal & 30.0 & 46.7 & \textbf{80.0} & --- \\
ChipsCup    & Nominal & 33.3 & 46.7 & \textbf{73.3} & --- \\
\bottomrule
\end{tabular}
\caption{Hardware success rates (\%) by task and perturbation condition.}
\label{tab:hw-sr-by-task}
\end{table}

\subsection{Additional Results}
\para{Inference Time Comparison}
We measure the inference latency of the frozen VLA and the steered closed-loop system over 50 replanning steps in simulation on a single NVIDIA A6000 Ada GPU. Table~\ref{tab:inference-time} reports the component-wise latency and the resulting effective replanning frequency.

The frozen VLA replans at 14.7 Hz. Adding $\langpolicy$ reduces the replanning frequency to 3.6 Hz, with most of the added latency coming from autoregressive language generation rather than VLA inference. Since each replan step corresponds to 5 environment steps at 20 Hz, the steered system still operates at a closed-loop update rate suitable for our experiments, but language generation is the primary computational bottleneck. This suggests that future implementations could improve latency through distilling $\langpolicy$, caching repeated language outputs, or updating language asynchronously rather than at every VLA query.

\begin{table}[H]
\centering
\small
\begin{tabular}{lcccc}
\toprule
\textbf{Config} & \textbf{$\vla$ (ms)} & \textbf{$\langpolicy$ (ms)} & \textbf{Total (ms)} & \textbf{Replan Hz} \\
\midrule
$\vla$ only & $68$ & --- & $68$ & $14.7$ \\
$\vla \circ \langpolicy$ & $70$ & $204$ & $274$ & $3.6$ \\
\bottomrule
\end{tabular}
\caption{Inference latency per replanning step, measured over 50 replanning steps on a single NVIDIA A6000 Ada GPU. Each replanning step corresponds to 5 environment steps at 20\,Hz.}
\label{tab:inference-time}
\end{table}

\para{Successful Trajectory Length}
We also compare the length of successful hardware rollouts in the nominal setting. Table~\ref{tab:nominal_mean_traj_len} reports the mean number of environment steps among successful trajectories only. Across most tasks, $\langpolicy$ completes the task in fewer steps than the base VLA, suggesting that language steering not only improves success but can also elicit more direct execution strategies. The largest reductions appear on \textbf{MarkerBlock}, \textbf{Microwave}, and \textbf{ChipsCup}, where $\langpolicy$ shortens successful trajectories by 23.1\%, 14.8\%, and 35.4\% relative to $\vla$, respectively.

However, shorter trajectories are not universal: $\sftpolicy$ is fastest on \textbf{CubeSort} and \textbf{MarkerBowl}. Together with the success-rate results, these trajectories indicate that interactive language search can discover language that both improves reliability and, in many cases, induces more efficient execution.

\begin{table}[H]
  \centering
  \small
  \begin{tabular}{@{}lccc@{}}
    \toprule
    Task & $\vla$ & $\sftpolicy$ & $\langpolicy$ \\
    \midrule
    \textbf{CubeSort} & 709.5 & \textbf{587.5} & 627.4 \\
    \textbf{CubeMug} & 680.9 & 578.3 & \textbf{542.1} \\
    \textbf{Microwave} & 499.0 & 469.2 & \textbf{425.2} \\
    \textbf{MarkerBlock} & 775.6 & 696.1 & \textbf{596.7} \\
    \textbf{MarkerBowl} & 631.5 & \textbf{578.1} & 657.0 \\
    \textbf{ChipsCup} & 870.8 & 728.4 & \textbf{562.2} \\
    \bottomrule
  \end{tabular}
  \caption{Mean successful trajectory length (steps) on nominal hardware tasks.}
  \label{tab:nominal_mean_traj_len}
\end{table}

\para{Ablation: Narration Data Size}
We ablate $\sftpolicy$ training with 10, 30, 50 narration videos per task; for each, we train $\langpolicy$ and deploy in Visual-Viewpoint (5 semantic perturbs, 50 rollouts/combo). The table below shows SFT data size has minimal effect on final closed-loop performance; the main gains come from the RFT phase involving \textit{interaction} with $\vla$ rather than from the narration phase. 

% , indicating the pipeline is robust to narration data size.

\begin{table}[h!]
\vspace{-1em}
\centering
\small
\setlength{\tabcolsep}{4pt}
\renewcommand{\arraystretch}{0.8}
\begin{tabular}{l|c|ccccc}
\toprule
\textbf{Model (Data)} & $\vla$  & $\langpolicy$ (10) & $\langpolicy$ (30) & $\langpolicy$ (50) \\
\midrule
\textbf{SR(\%)} & 51.2 $\pm$ 1.0 & 60.8 $\pm$ 1.0 & 59.7 $\pm$ 1.0 & 63.9 $\pm$ 1.0 \\
\bottomrule
\end{tabular}
\caption{Narration data size ablation on Visual-Viewpoint tasks.}
\label{tab:nar_size_ablation}
\vspace{-1em}
\end{table}
\end{document}